%% file: main.tex
\documentclass[10pt,twocolumn,letterpaper]{article}

\usepackage{iccv}
\usepackage{times}
\usepackage{epsfig}
\usepackage{graphicx}

\usepackage{capt-of}
\usepackage{xcolor}
\usepackage{amsmath,amssymb,amsfonts,dsfont,pifont,bm,bbm,mathrsfs,mathtools,nicefrac}
\usepackage[linesnumbered,ruled,algo2e]{algorithm2e}
\usepackage{algorithm,listings}
\usepackage{algorithmicx}
\usepackage[noend]{algpseudocode} 

\floatname{algorithm}{Algorithm}

\usepackage{booktabs,multirow,adjustbox,diagbox,threeparttable}
\definecolor{citeblue}{RGB}{48,111,186}
\usepackage[caption=false,textfont=footnotesize,labelfont=footnotesize,position=bottom]{subfig}
\usepackage[pagebackref=true,breaklinks=true,colorlinks,bookmarks=false,citecolor=citeblue,]{hyperref}
\usepackage[capitalize]{cleveref}  

\usepackage{titletoc}
\newcommand\DoToC{%
  \startcontents
  \printcontents{}{1}{\hrulefill\vskip0pt}
  \vskip0pt \noindent\hrulefill
  }

\crefname{section}{Sec.}{Secs.}
\Crefname{section}{Section}{Sections}
\crefname{table}{Tab.}{Tabs.}
\Crefname{table}{Table}{Tables}
\crefname{figure}{Fig.}{Figs.}
\Crefname{figure}{Figure}{Figures}
\crefname{equation}{Eq.}{Eqs.}
\Crefname{equation}{Equation}{Equations}
\newcommand{\best}{\textcolor[rgb]{ .875,  .165,  .247}}
\newcommand{\seco}{\textcolor[rgb]{ .78,  .361,  0}}

\newcommand{\tocite}[1]{\textcolor{red}{[TO CITE]}}

\newcommand{\para}{\bm{\Theta}}
\newcommand{\ce}{\gL_{\rm ce}}
\newcommand{\gkl}{\nabla \mathcal{L}_{\text{kl}}}
\newcommand{\gce}{\nabla \mathcal{L}_{\text{ce}}}

\newcommand{\std}[1]{\scriptsize{$\pm \emph{#1}$}}

\iccvfinalcopy 

\def\iccvPaperID{2258} 

\ificcvfinal\fi
\newcommand\nonumfootnote[1]{%
\begingroup%
    \renewcommand\thefootnote{}\footnote{\hspace{-3.7pt}#1}%
    \addtocounter{footnote}{-1}%
\endgroup%
}

\input{math_cmd.tex}

\begin{document}

\title{Regularized Mask Tuning: Uncovering Hidden Knowledge \\in Pre-trained Vision-Language Models}

\author{
Kecheng Zheng$^{\dagger,1,2}$ \quad Wei Wu$^{\dagger,4}$ \quad Ruili Feng$^{4}$ \quad Kai Zhu$^{4}$ \quad Jiawei Liu$^{4}$ \\ Deli Zhao$^{3}$ \quad Zheng-Jun Zha$^{4}$ \quad Wei Chen$^{1}$ \quad Yujun Shen$^2$
\\[0.1025cm]
\textsuperscript{1}State Key Lab of CAD\&CG, Zhejiang University\quad \textsuperscript{2}Ant Group\quad\textsuperscript{3}Alibaba Group
\quad\textsuperscript{4}USTC
  }

\maketitle

\input{sections/0.abs.tex}
\input{sections/1.intro.tex}

\input{sections/2.related.tex}
\input{sections/3.method.tex}

\input{sections/4.exp.tex}
\input{sections/5.conclusion.tex}
\input{sections/6.ref.tex}
\input{sections/appendix.tex}

\end{document}

%% file: math_cmd.tex
\usepackage{amsmath,amsfonts,bm}

\def\eqref#1{equation~\ref{#1}}

\def\1{\bm{1}}

\def\vtheta{{\bm{\theta}}}

\def\vb{{\bm{b}}}

\def\vx{{\bm{x}}}
\def\vy{{\bm{y}}}

\def\mG{{\bm{G}}}

\def\mM{{\bm{M}}}

\DeclareMathAlphabet{\mathsfit}{\encodingdefault}{\sfdefault}{m}{sl}
\SetMathAlphabet{\mathsfit}{bold}{\encodingdefault}{\sfdefault}{bx}{n}

\def\gL{{\mathcal{L}}}
\def\gM{{\mathcal{M}}}
\def\gN{{\mathcal{N}}}

\def\sR{{\mathbb{R}}}



%% file: sections/0.abs.tex
\begin{abstract}

Prompt tuning and adapter tuning have shown great potential in transferring pre-trained vision-language models (VLMs) to various downstream tasks.
In this work, we design a new type of tuning method, termed as \textbf{regularized mask tuning}, which masks the network parameters through a learnable selection.
Inspired by neural pathways, we argue that the knowledge required by a downstream task already exists in the pre-trained weights but just gets concealed in the upstream pre-training stage.
To bring the useful knowledge back into light, we first identify a set of parameters that are important to a given downstream task, then attach a binary mask to each parameter, and finally optimize these masks on the downstream data with the parameters frozen.
When updating the mask, we introduce a novel gradient dropout strategy to regularize the parameter selection, in order to prevent the model from forgetting old knowledge and overfitting the downstream data.
Experimental results on 11 datasets demonstrate the consistent superiority of our method over previous alternatives.
It is noteworthy that we manage to deliver \textbf{18.73\%} performance improvement compared to the zero-shot CLIP via masking an average of only \textbf{2.56\%} parameters.
Furthermore, our method is synergistic with most existing parameter-efficient tuning methods and can boost the performance on top of them.
Project page can be found \href{https://wuw2019.github.io/R-AMT/}{here}.
\nonumfootnote{$\dagger$ indicates equal contribution.}

\end{abstract}

%% file: sections/1.intro.tex
\section{Introduction}\label{sec:intro}

\definecolor{flamecolor}{RGB}{233,161,72}
\definecolor{snowflakecolor}{RGB}{91,157,219}
\begin{figure}
    \centering
    \includegraphics[width=1.0\linewidth]{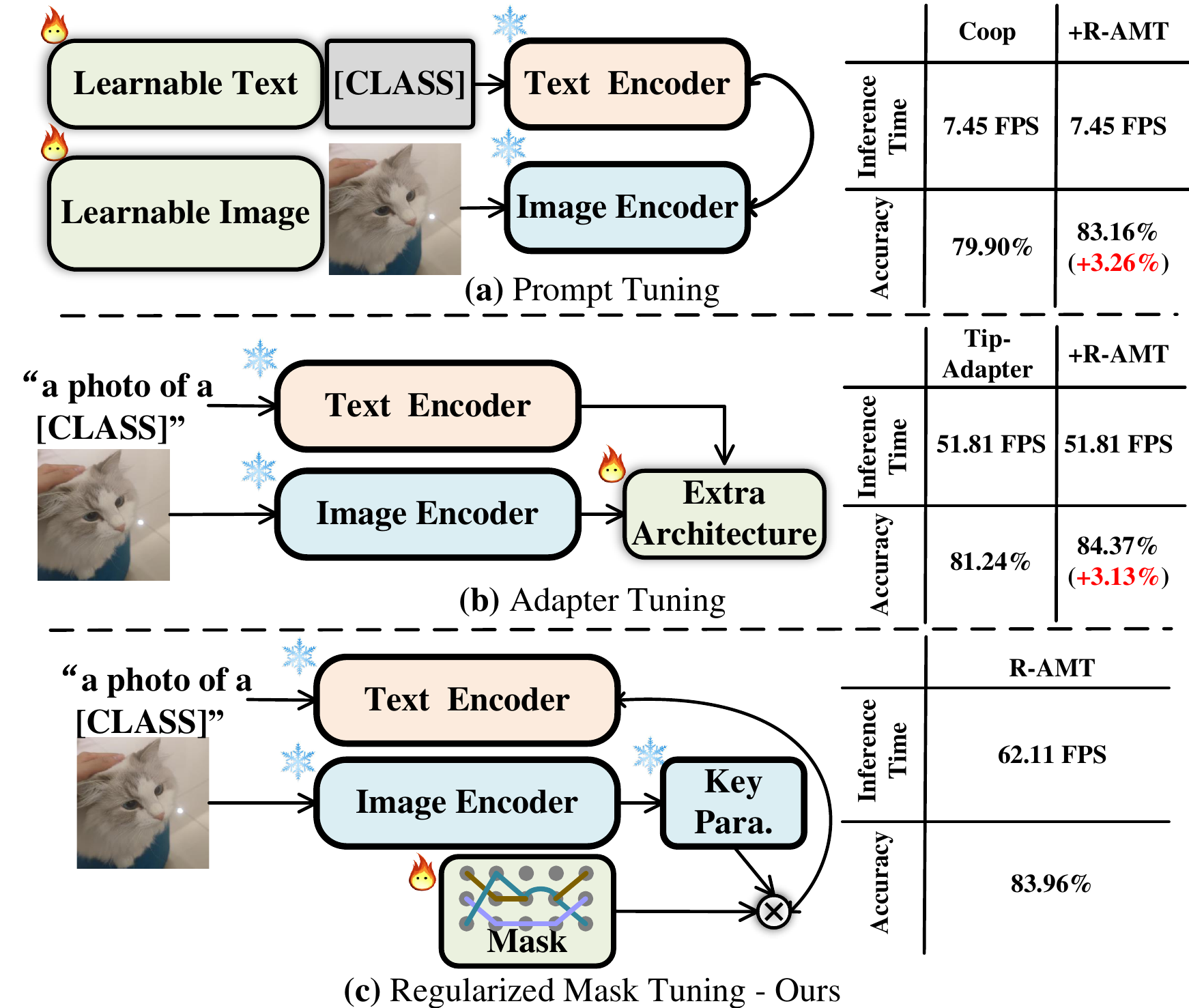}
    \caption{%
        \textbf{Concept diagrams} of (a) prompt tuning~\cite{zhou2022learning,zang2022unified}, (b) adapter tuning~\cite{gao2021clip,zhang2022tip}, and (c) our \textit{regularized mask tuning}. 
        The tables on the right of (a)(b) demonstrate the inference time and accuracy of the existing tuning method before and after combining with our regularized mask tuning method (R-AMT). The R-AMT significantly boosts their performance without introducing additional inference time.
        ``Key Para.'' refers to the identified key parameters (\eg, multi-head self-attetnion). \textbf{\textcolor{flamecolor}{Flames}} and \textbf{\textcolor{snowflakecolor}{snowflakes}} refer to learnable and frozen parameters, respectively.
    }
    \label{fig:moti}
\end{figure}

The advent of large-scale pre-trained vision-language models (VLMs)~\cite{radford2021learning} has ushered in a new era of incorporating language features to supervise the image encoder for a wide range of downstream visual tasks, such as few-shot learning~\cite{zhou2022learning} and open-world detection~\cite{du2022learning}. 
Thanks to the multimodal architecture and millions of text-image pairs from the web, VLMs exhibit exceptional zero-shot transferability in downstream tasks. 
To further enhance the transferability of VLMs, researchers have proposed efficient tuning methods, such as adapter tuning~\cite{gao2021clip,zhang2022tip} or prompt tuning~\cite{zhou2022learning,zang2022unified,lu2022prompt}. 
These techniques incorporate a small number of task-specific parameters and train them solely on the downstream task, thus significantly improving the performance and reducing computational requirements.

The essence of efficient tuning methods lies in two fundamental components, \ie leveraging the well-learned knowledge structure of VLMs and efficiently exploring the task-specific knowledge given few-shot data. 
Despite its potential, however, most existing efficient transfer learning approaches direct utilize all parameters of pre-trained VLMs and do not consider further unleashing the potential of the well-learned knowledge of VLMs. 
Specifically, prompt tuning methods~\cite{zhou2022learning} use the frozen CLIP model and add the extra learnable parameters from the input side as shown in \cref{fig:moti}a. 
Adapter modules~\cite{gao2021clip,zhang2022tip} consist of a small set of the learnable module, further inserted into the frozen pre-trained model for adaptation as in~\cref{fig:moti}b.
Despite the considerable efforts in efficient tuning methods from the prompt or adapter side, these methods do not explore the frozen CLIP parameters at all, choosing instead to add additional modules to learn task-specific knowledge.
Thus, as shown in~\cref{fig:moti}c, we adopt \textbf{\textit{mask tuning}} to explore the well-learned knowledge structure of VLMs and uncover the hidden knowledge in them for task-specific domains.

In the field of neural physiology~\cite{
hubel1962receptive,engel1997retinotopic,zeki1988functional}, it has been discovered that neurons in the brain cortex exhibit diverse knowledge of various visual features such as shape, color, and depth. 
The knowledge is distributed in distinct neurons that have specific functions and work in conjunction with one another, termed neural pathways.
When there is knowledge of a new environment coming, the neurons will compare it with the old knowledge learned in the past and then pick new conjunctions (\ie, neural pathways) to adapt to the new environment.
Analogous to VLMs, parameters act as a manifestation of neurons and are responsible for memorizing knowledge from data. Thus, selecting suitable parameters as parameter pathways is beneficial for uncovering the key knowledge of downstream tasks.

Inspired by the neural pathways, we propose an efficient \textit{Regularized Mask Tuning (R-MT)} method to mask the parameters of the pre-trained VLMs under a learnable selection.
Specifically, we first identify a subset of the parameters (\eg, multi-head self-attentive layer) based on the magnitude of the gradient changes as sensitive network parameters for downstream tasks.
Then, we introduce a binary mask equipped with gradient dropout regularization to the selected parameters.
Because few-shot training tends to cause overfitting, we introduce the logits from pre-trained VLMs as the general knowledge to prevent mask tuning from forgetting.
Concretely, the gradient dropout regularity as an effective regularizer introduces the probabilistic masking strategy that samples gradients based on the level of consistency of the downstream-related knowledge and the general knowledge, which can reject weak loss minima that may lead to overfitting.
Our findings indicate that selecting well-placed parameters is crucial for achieving successful transfer settings.
Moreover, our method is orthogonal to most existing parameter-efficient adaption methods (\eg, adapter and prompt) and endows them the ability to customization on downstream needs.
Extensive experiments on 11 datasets demonstrate the effectiveness of the proposed method.

%% file: sections/2.related.tex
\section{Related Work}\label{sec:related-work}

\noindent\textbf{Vision-language models} achieve cross-modality alignment by learning a joint embedding space for text and image representation. A typical VLM consists of three components: text encoder, image encoder, and alignment function. The text and image encoder is trained separately before being connected by the alignment function in the early stage~\cite{frome2013devise}.
Recent VLMs such as CLIP \cite{radford2021learning} and Align~\cite{jia2021scaling} jointly optimize text and image encoder through contrastive learning. Benefiting from the millions of text-image pairs from the web and the multi-modality structure, these VLMs achieve exceptional zero-shot transfer capacity in the downstream tasks. Toward better transfer ability, researchers propose a series of parameter-efficient methods to adapt CLIP to downstream tasks, such as image recognition~\cite{zhou2022learning,zang2022unified,gao2021clip,zhang2022tip,guo2022calip}.

\noindent\textbf{Parameter-efficient adaption methods} for CLIP can be coarsely divided into two categories: prompt tuning~\cite{zhou2022learning,zhang2023prompt,zang2022unified,chen2022prompt} and adapter tuning~\cite{gao2021clip,zhang2022tip}. 
Inspired by the success of prompt learning in NLP~\cite{brown2020language,li2021prefix,gao2020making}, some researchers involve prompt learning methods in CLIP to improve the few-shot transfer capacity. Zhou \etal~\cite{zhou2022learning} first introduce learnable text prompts to adapt CLIP to visual recognition tasks, which brings a great improvement over Zero-shot CLIP. Zang \etal~\cite{zang2022unified} propose a unified prompt learning strategy for text and image encoder, which simultaneously refine the text and image representation for adapting to downstream tasks.
Adapter modules consist of a small set of learnable parameters, which are further inserted into the frozen pre-trained model for adaptation. Gao \etal~\cite{gao2021clip} add adapters after text and image branches through residual connection. Zhang \etal~\cite{zhang2022tip} employ a training-free adapter module following the image encoder, which is initialized using the knowledge extracted from the downstream training set. 
However, existing methods mainly focus on changing the input of CLIP (\ie, text prompt tuning~\cite{zhou2022learning} and visual prompt tuning~\cite{zang2022unified}) or adding extra modules out of CLIP (\ie, adapter tuning~\cite{gao2021clip,zhang2022tip}), which neglects to excavate the inner power of CLIP.

\noindent\textbf{Binary mask} is commonly used to find a subnetwork structure from the model, which can be viewed as a way of network pruning. It can be achieved through a straight-through estimator~\cite{bengio2013estimating,rastegari2016xnor}.
Csordás \etal~\cite{csordas2020neural} learn binary masks to identify subnetworks responsible for different tasks. Zhang \etal~\cite{zhang2021can} search subnetworks with binary mask to achieve better out-of-distribution (OOD) performance. These works focus on finding a functional subpart of weights inside a given pre-trained neural network, which can be retrained for new tasks. 
However, Zhou~\etal~\cite{zhou2019deconstructing} find that applying binary mask with a model is also a way to train the model by investigating the lottery ticket hypothesis of network pruning.  Recently, researchers~\cite{liu2018rethinking,zhao2020masking} propose that training binary mask for a pre-trained language model is similar to finetuning and is more parameter-efficient.
Moreover, Mallya \etal~\cite{mallya2018piggyback} train binary mask with fixed convolutional neural network for image classification, which achieves good performance. These works demonstrate the capacity of binary masks in parameter-efficient training in natural language processing and computer vision. 
Different from these methods, we propose a regularized mask tuning to search an important subset of weights in the image encoder of fixed CLIP for downstream tasks. Moreover, the regularized mask tuning can be further combined with other parameter-efficient methods presuming better few-shot performance.

%% file: sections/3.method.tex
\section{Method}\label{sec:method}

\begin{figure*}[t]
    \centering
    \includegraphics[width=0.85\linewidth]{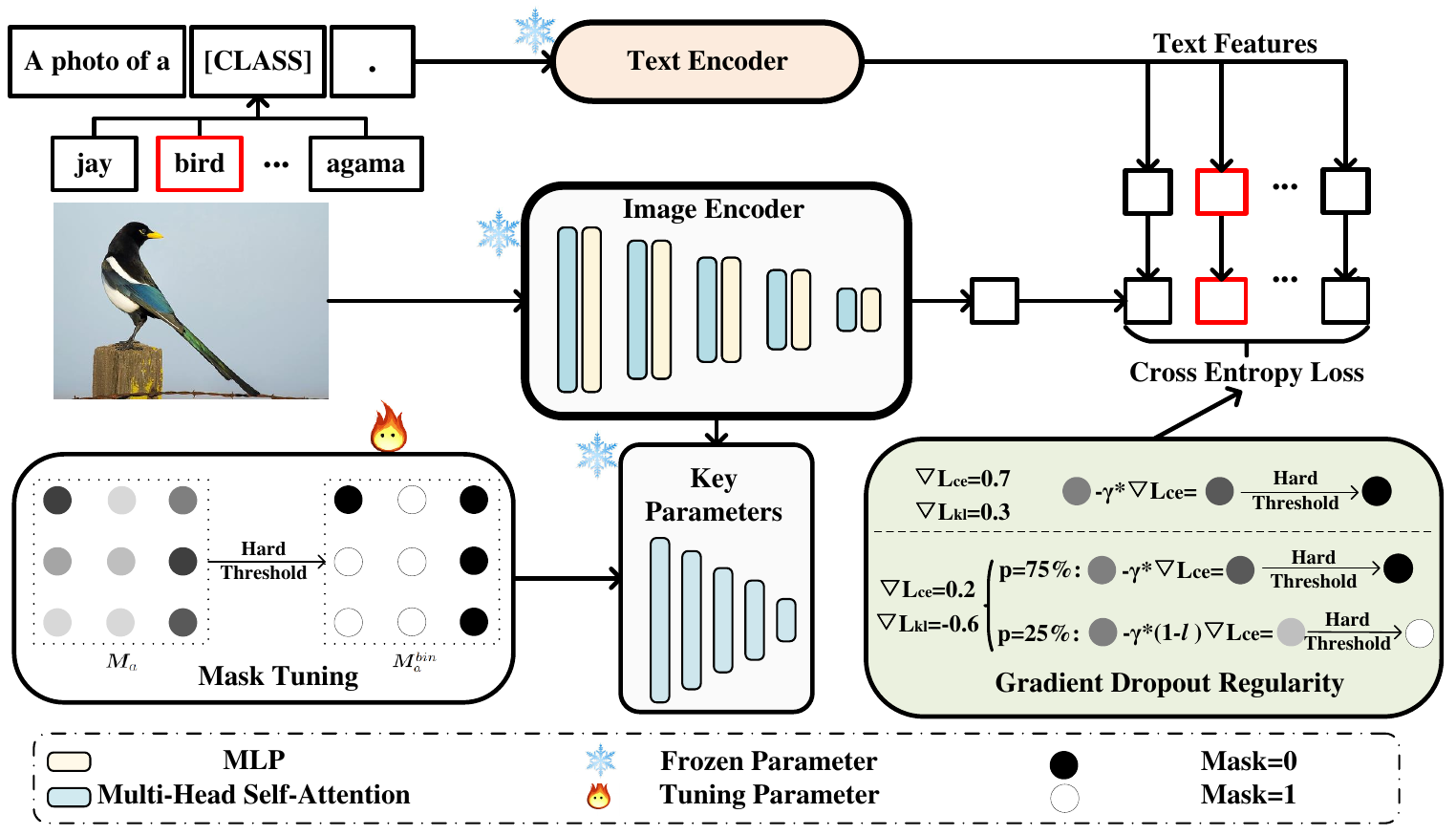}
    \caption{%
        Overview of the proposed regularized mask tuning for frozen CLIP. During training, we maintain a set of mask weights $\textbf{M}_a$ which are passed through a threshold function to obtain binary masks $\textbf{M}^{bin}_a$. There, we select the MHSA as the key parameter. When updating the mask, we introduce a novel gradient dropout strategy to regularize the parameter selection, in order to prevent the model from forgetting general knowledge from pre-trained CLIP and overfitting the downstream data.
        %
    }
    \vspace{-5pt}
    \label{fig:arch}
\end{figure*}

In this section, we introduce the Regularized Mask
Tuning (RMT) method in detail, which aims to better adapt CLIP to downstream tasks.  
%
%

\subsection{Preliminaries of CLIP}
\label{sec:method_clip}
CLIP~\cite{radford2021learning} mainly consists of two components: image encoder $\mG_{I}(\vtheta)$ and text encoder $\mG_{T}(\bm{\beta})$, which are designed to project image and text into the same feature embedding space. 
Specifically, given the images $\{\boldsymbol{x}_1, \boldsymbol{x}_2,\cdots,\boldsymbol{x}_m\}$ and a set of corresponding categories, the image recognition task aims to classify an image to a specific category. Here, the $m$ denotes the number of images in the dataset. 
To zero-shot adapt CLIP to the image recognition task, the name of category $\boldsymbol{y}_i$ is filled into a set of words, \textit{e.g.} ``a photo of a [class]'', to construct a hand-craft text prompt $\boldsymbol{t}_i$ as the input of text encoder. 
The possibility of an image $\boldsymbol{x}_j$ being assigned to class $\boldsymbol{y}_i$ is formulated as following:
\begin{gather}
    \boldsymbol{g}_i = G_T(\boldsymbol{t}_i;\vtheta), \boldsymbol{f}_j = G_I(\boldsymbol{x}_j;\bm{\beta}), \\
    p(\boldsymbol{y}=i \mid \boldsymbol{x}_j)=\frac{\exp \left(\cos \left(\boldsymbol{g}_i, \boldsymbol{f}_j\right) / \tau\right)}{\sum_{n=1}^k \exp \left(\cos \left(\boldsymbol{g}_n, \boldsymbol{f}_j\right) / \tau\right)},
    \label{eq.possibility}
\end{gather}
where the $\text{cos}(\cdot, \cdot)$ denotes the cosine similarity between two inputs and $\tau$ is a learnable temperature parameter.

\subsection{Mask Tuning}
Although CLIP has strong zero-shot performance, we argue that the knowledge required by a downstream task already exists in the pre-trained weights but may get concealed by some unnecessary information emerging in the upstream pre-training. To uncover the hidden valuable knowledge in pre-trained weights, we aim to identify the weights required by a downstream task, termed as a neural pathway, to facilitate few-shot learning~\cite{zhou2022learning}. 
Concretely, we take the parameters $\vtheta$ of the image encoder as an example for analysis. Given $\vtheta=(\vtheta_1,\cdots,\vtheta_n)^T\in\para\subset\sR^N$ where $N$ is the parameter volume, our aim is to learn a binary mask matrix $ \mM^{bin}$ as a downstream-related neural pathway:
\begin{align}
    \label{eq.weight_matrix}
    \vtheta_{\gM} := \vtheta\odot \mM^{bin}, & 
\end{align}
where the $\odot$ refers to Hadamard product and $\vtheta_{\gN}^T\in\para_{\gN}\subset\sR^n, n\ll N,$ refers to a small subset of pre-trained weights. By utilizing solely the parameters of the subset of pre-trained weights, it is adequate to transfer to the downstream domain. Since the binarized function shown in \cref{eq.weight_matrix} is non-differentiable, we conduct a real-valued mask weight $\boldsymbol{M}$ and pass it through an element-wise thresholding binary function to obtain
$\boldsymbol{M}^{bin}$. Meanwhile, we use the gradient $\frac{\partial \mathcal{L}_{\text{ce}}}{\partial\boldsymbol{M}^{bin}}$ as a noisy estimator of $\frac{\partial \mathcal{L}_{\text{ce}}}{\partial\boldsymbol{M}}$ to update the $\boldsymbol{M}$, following the previous work \cite{zhao2020masking,lin2020dynamic}. The optimization can be formulated as:
\begin{equation}
    \label{eq.optimize}
    \boldsymbol{M} \gets \boldsymbol{M}-\gamma \frac{\partial \mathcal{L}_{\text{ce}}}{\partial\boldsymbol{M}^{bin}},
\end{equation}
where the $\gamma$ denotes the learning rate that controls the sparsity of mask, and $\mathcal{L}_{\text{ce}}$ denotes the Cross-Entropy (CE) loss. The binary mask $\mM^{bin}=\mathcal{I}[\mM>\alpha]$, where $\alpha$ is a hard threshold.
An astonishing discovery is that setting \textbf{0.16\%}
parameters of CLIP image encoder to 0 results in a performance improvement of \textbf{44.40\%} compared to the zero-shot performance in the EuroSAT. This discovery supports the notion that certain parameters contain valuable knowledge for downstream tasks, which are also duplicated in redundant parameters. Consequently, selecting an efficient neural pathway from pre-trained weights significantly influences performance.


\paragraph{Which layers to apply binary mask?} 

\begin{figure}[htbp]
    \centering
    \includegraphics[width=0.82\linewidth]{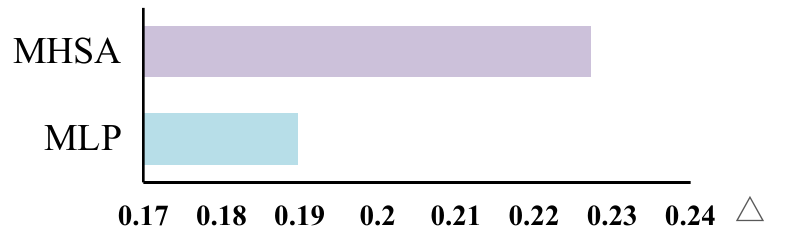}
    \caption{%
        Analysis of change in mask weights $\mM$ when fine-tuning to downstream tasks with $\gce$. Over 11 datasets, the mean change in the MHSA layers is significantly higher than MLP over.
    }
    \label{fig:change}
    \vspace{-0.1in}
\end{figure}
While this method is capable of efficiently identifying the parameter pathway that is most suitable for the downstream task, the sheer number of mask parameters to be trained can be overwhelming. As a result, it is crucial to devise a means of assessing parameter importance, by identifying the relevant neural pathway based on these significant parameters to mask. This method represents a more balanced approach, one that strikes a delicate balance between computational effort and overall performance. Our goal is to identify a subset of weights that can be effectively transferred to the downstream task while retaining important general information about the model. 

To achieve this, we analyze the change in mask weight $\mM$ for each layer after training on the target dataset with the CE loss, \ie, $\Delta = \sum \gamma* \frac{\partial\ce}{\partial\mM}$.
These parameters come from two types of layers -- (1) multi-head self-attention (MSHA) layers and (2) multilayer perception (MLP) layers. We present the mean $\Delta$ when the mask weight is training on the 11 datasets in~\cref{fig:change}.
As we see, the MSHA layer parameters have relatively higher $\Delta$ compared to the MLP layer. Moreover, MSHA layers are $20\%$ of the total parameter count in the model and achieve the same performance (\eg, 83.96\% vs. 83.96\% on average over 11 datasets). This suggests binary attention mask $\boldsymbol{M}^{bin}_a$ plays a significant role during fine-tuning, and we leverage that in our method as shown in~\cref{fig:arch}. We name this method \textbf{Attention Mask Tuning} (AMT). Moreover, we term binary mask on all layers as Parameter Mask Tuning (PMT) and MLP layers as Multilayer perception Mask Tuning (MMT), respectively, for distinction.

\subsection{Gradient Dropout Regularity}

Stochastic Gradient Descent (SGD)~\cite{robbins1951stochastic} is a popular optimization algorithm used in machine learning to minimize a loss function during training. SGD works by randomly selecting a small subset of training examples to compute the gradient of the loss function. It can help to avoid overfitting, as it adds some level of randomness to the gradient updates. This helps to prevent the algorithm from getting stuck in local minima and encourages exploration of the solution space. But in few-shot learning scenarios, particularly in 1-shot or 2-shot learning, the mini-batch data is typically derived from the entire training set to compute the gradient. Thus, this approach lacks the stochastic property of traditional SGD, which can lead to the overfitting of the model to the training data.

In order to make our binary tuning method better suited for few-shot scenarios, we develop Gradient Dropout Regularity formalism that randomly introduces the gradient regularity to reduce the amount of overfitting that occurs and help the model generalize better to new domains. 
We deem the zero-shot CLIP predictions as the general knowledge and the label from the downstream task as the target-specific knowledge.
Then we introduce the Kullback-Leibler (KL) divergence between them to regularize the gradient.
To implement Gradient Dropout Regularity, we first define the Gradient Retaining Purity $\mathcal{P}$ as follows 
\begin{equation} 
\label{eq:psp} 
\mathcal{P} = \frac{1}{2}\left(1+\frac{\text{sgn}(\gce)\left(\gce + \gkl\right)}{|\gce| + |\gkl|}\right),
\end{equation}
where $\mathcal{P}$ is bounded by $[0,1]$. There are two ways to describe the relationship between $\gce$ and $\gkl$. 
Firstly, their sign is the same target-specific, implying that the optimization direction of few-shot downstream knowledge is compatible with general knowledge. Thus, we can safely update the gradient as $\gce$.
Secondly, their signs are different at the updated position, indicating that optimizing the binary mask with $\gce$ will result in the loss of pre-trained general knowledge. This implies that few-shot downstream knowledge conflicts with general knowledge. 
In this case, we regularize the $\gce$ via random gradient dropout strategy under the guidance of $\gkl$ to optimize the model for classification. We mathematically rewrite the~\cref{eq:psp} formulated as:
\begin{equation}\label{eq:projgrad}
\mathcal{P}= 
\left\{
             \begin{array}{ll}
             1, & \text{if}\ \gce \cdot\gkl \geq 0 \\
             (1+\frac{\gce + \gkl}{|\gce| + |\gkl|})/2,& \text{if}\ \gce>0\ \text{and}\ \gkl<0 \\
             (1-\frac{\gce + \gkl}{|\gce| + |\gkl|})/2,& \text{if}\ \gce<0\ \text{and}\ \gkl>0.
             \end{array}
\right.
\end{equation}
Thus, $\mathcal{P}$ is a measure of the agreement of general and target-specific knowledge at the updated position.
We the formulate a gradient dropout function $\mathcal{M}_{\text{ce}}$ as: $\mathcal{M}_{\text{ce}} = \mathcal{I}[\mathcal{P}>U]$,
where $\mathcal{I}$ the standard indicator function. 
$U$ is a tensor composed of i.i.d $U(0,1)$ random variables. 
The optimization can be formulated as:
\begin{equation}
    \label{eq.optimize_final}
    \boldsymbol{M}_a \gets  \boldsymbol{M}_a-\gamma * (1-l+l*\mathcal{I}[\mathcal{P}>U]) * \frac{\partial \mathcal{L}_{\text{ce}}}{\partial\boldsymbol{M}_a^{bin}},
\end{equation}
where $l \in [0,1]$ is a leak parameter. $l<1$ means we allow $\gce$ leak through.
The complete Gradient Dropout Regularity technique involves computing the purity metric $\mathcal{P}$ at each gradient point and building a gradient consistency framework for cross-entropy loss with the help of KL divergence guidance. The steps for this are outlined in~\cref{alg.PPL}. We name the AMT with the Gradient Dropout Regularity technique as R-AMT.
Similarly, PMT and MMT with the Gradient Dropout Regularity technique are named R-PMT and R-MMT, respectively.

\begin{algorithm}[t]
	\caption{Regularized Mask Tuning}
	\label{alg.PPL}
	\KwIn{The image encoder $\boldsymbol{G}_I$ and text encoder $\boldsymbol{G}_T$ of CLIP, data $\mathcal{D}_{train}$ for downstream task, hard threshold $\alpha$, and leak parameter $l$.}
	\KwResult{Mask $\boldsymbol{M}_a$ for image encoder}
	Construct hand-craft text prompt set $\mathcal{T}=\{\boldsymbol{t}_c\}^{\mathcal{C}}_{c=1}$ with the label set of $\mathcal{D}_{train}$
	
	Extract text features $\boldsymbol{g}_c=\boldsymbol{G}_T(\boldsymbol{t}_c), c=1,2,\cdots,\mathcal{C}$
	
	Initialize the learnable mask weight $\boldsymbol{M}_a$ according the the weight choices
	
	\For{$e \in [1, epoch]$}{
	Sample a mini-batch $\{\boldsymbol{x}_i,\boldsymbol{y}_i\}^{N}_{i=1}$ from $\mathcal{D}_{train}$
	
	Apply hard threshold $\alpha$ on the mask weight $\boldsymbol{M}_a$ to calculate CE loss $\mathcal{L}_{\text{ce}}$ and KL loss $\mathcal{L}_{\text{kl}}$
	
	
	

	



        Calculate $\mathcal{P} = \frac{1}{2}\left(1+\frac{\text{sgn}(\gce)\left(\gce + \gkl\right)}{|\gce| + |\gkl|}\right)$

        Sample $U$ from Uniform Distribution $U(0,1)$


        Set final gradient \\$\nabla_{final} = (1-l*(1-\mathcal{I}[\mathcal{P}>U]))\gce$
	
	Optimize the learnable matrix $\boldsymbol{M}_a$ with $\nabla_{final}$ by gradient descent:  $\boldsymbol{M}_a \gets  \boldsymbol{M}_a-\gamma \nabla_{final}$
	
	}
	
\end{algorithm}

%% file: sections/4.exp.tex
\section{Experiments}\label{sec:exp}
\subsection{Experimental Settings}

\noindent\textbf{Datasets}.
We conduct experiments on 11 publicly available image classification datasets following CoOP~\cite{zhou2022learning}. The datasets including ImageNet~\cite{deng2009imagenet}, FGVCAircraft~\cite{maji2013fine}, StanfordCars~\cite{Krause_2013_ICCV_Workshops}, Flowers102~\cite{nilsback2008automated}, Caltech101~\cite{fei2004learning}, DTD~\cite{cimpoi2014describing}, EuroSAT~\cite{helber2019eurosat}, Food101~\cite{bossard2014food}, UCF101~\cite{soomro2012ucf101}, OxfordPets~\cite{parkhi2012cats}, and SUN397~\cite{xiao2010sun}. 

\begin{figure*}[t]
\centering 
\subfloat[\textbf{Average over 11 datasets}]{
     \centering
     \includegraphics[width=0.30\linewidth,height=0.22\linewidth]{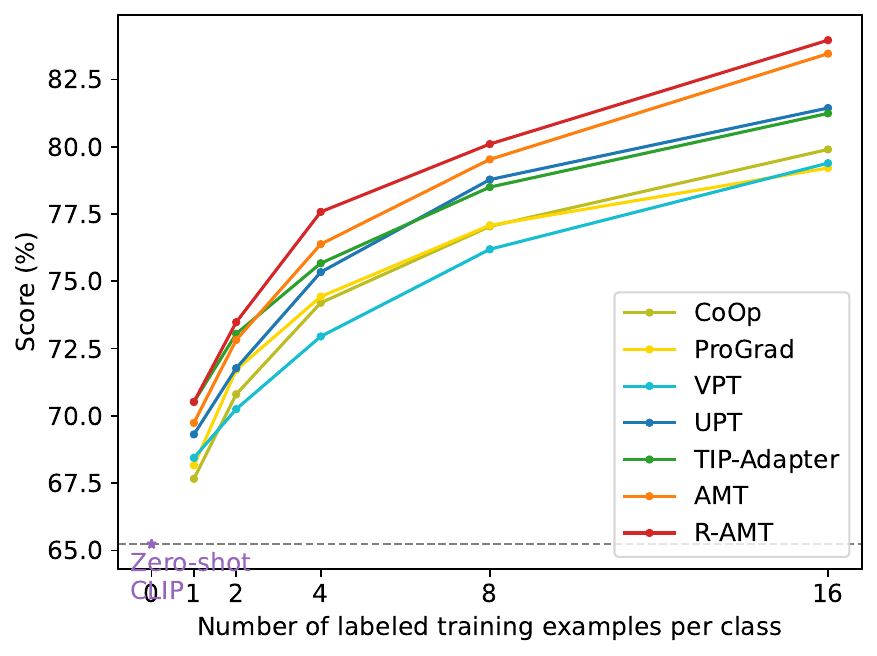}
     \label{fig:Average}
 }
 \hfill
\subfloat[FGVCAircraft]{
     \centering
     \includegraphics[width=0.30\linewidth,height=0.22\linewidth]{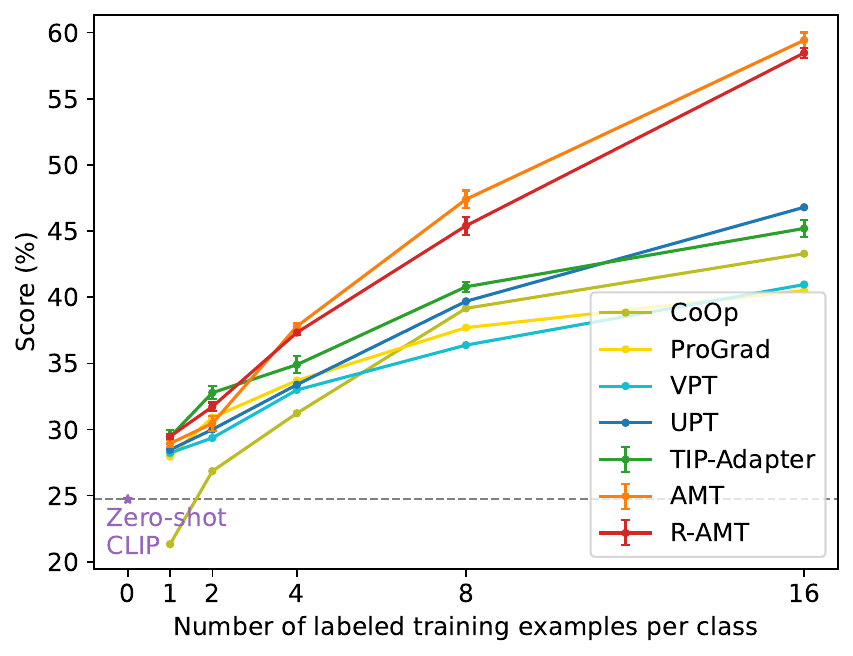}
     \label{fig:FGVCAircraft}
}
\hfill
\subfloat[ImageNet]{
     \centering
     \includegraphics[width=0.30\linewidth,height=0.22\linewidth]{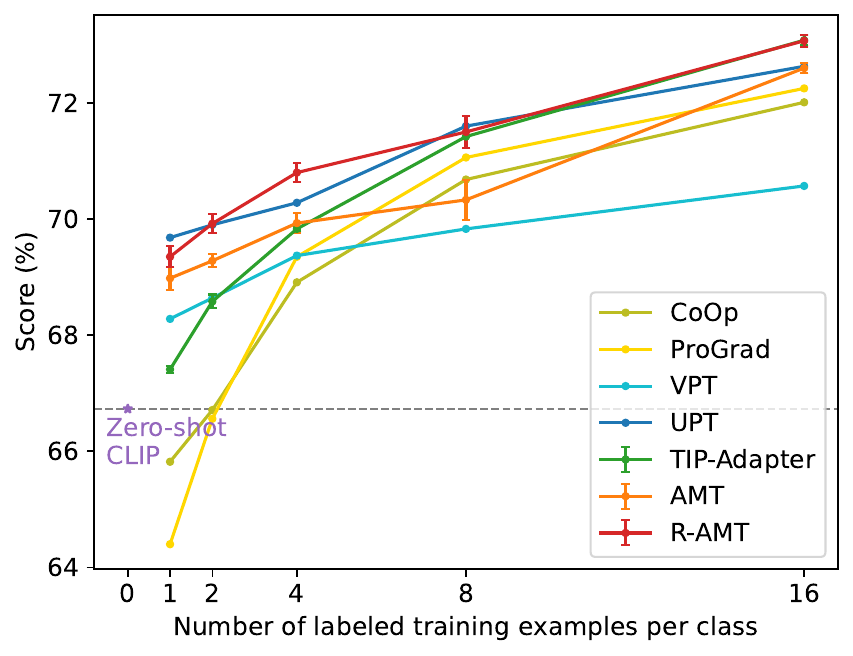}
     \label{fig:ImageNet}
}
\vspace{-10pt}
\subfloat[StanfordCars]{
     \centering
     \includegraphics[width=0.30\linewidth,height=0.22\linewidth]{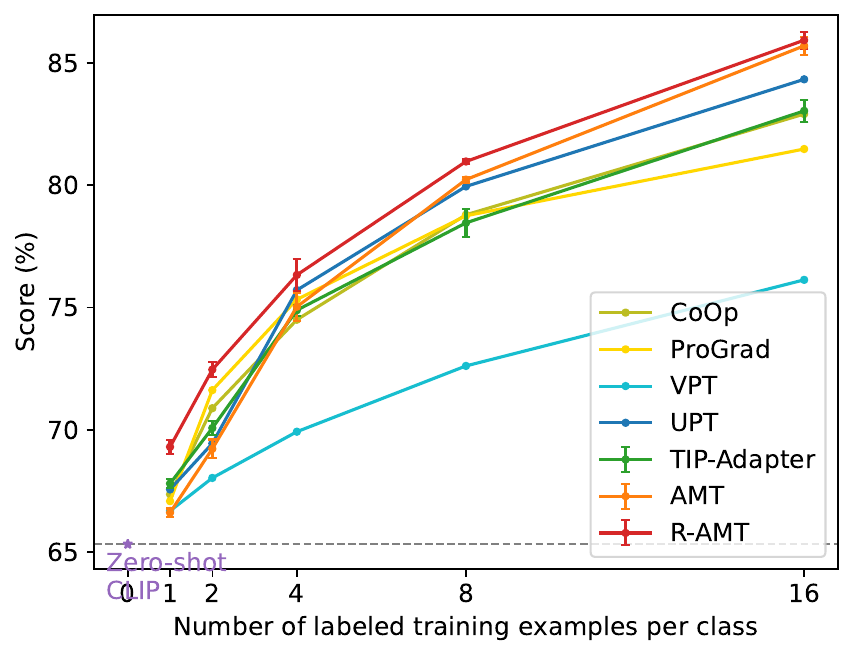}
     \label{fig:StanfordCars}
}
\hfill
\subfloat[Caltech101]{
     \centering
     \includegraphics[width=0.30\linewidth,height=0.22\linewidth]{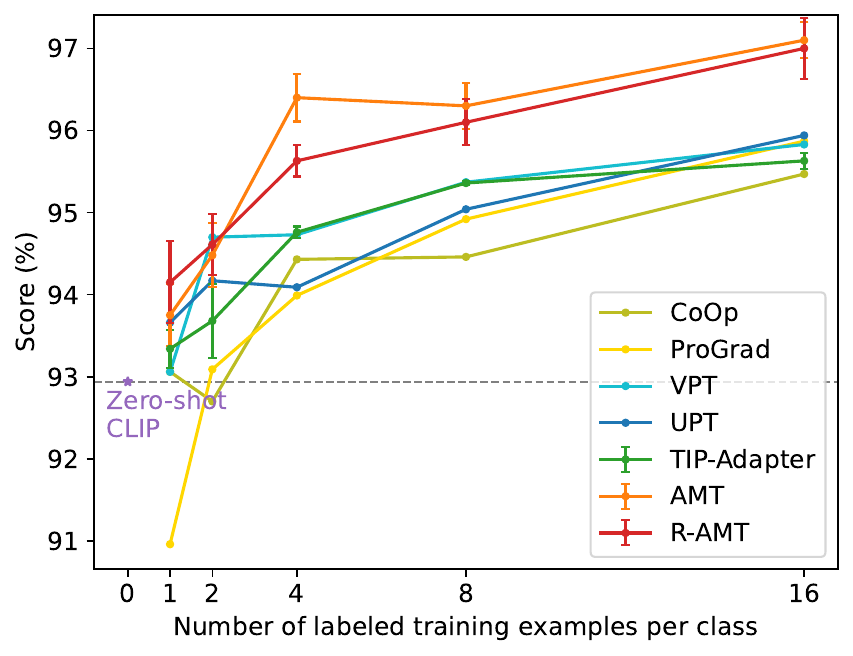}
     \label{fig:Caltech101}
}
\hfill
\subfloat[UCF101]{
     \centering
     \includegraphics[width=0.30\linewidth,height=0.22\linewidth]{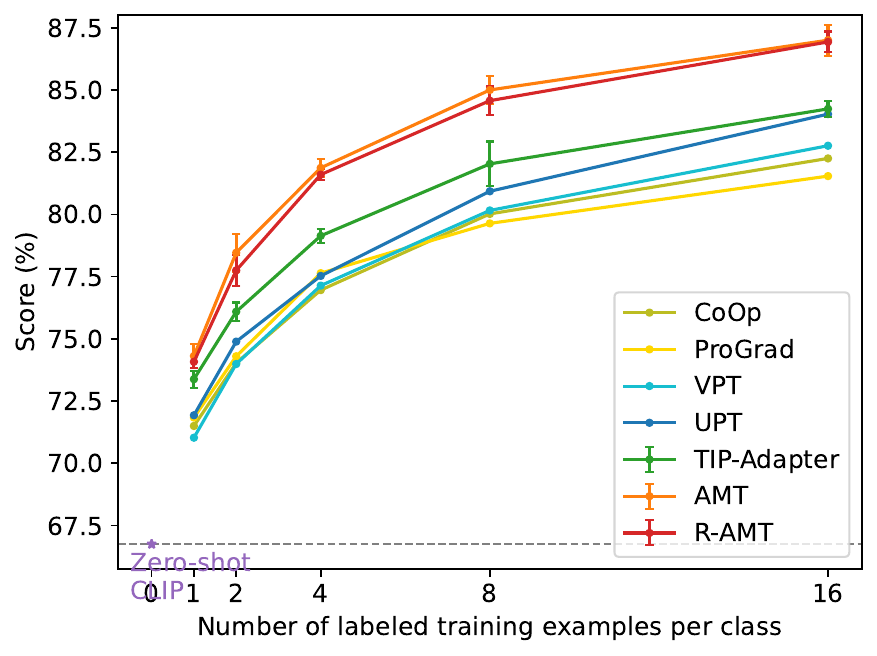}
     \label{fig:UCF101}
}
\vspace{-10pt}
\subfloat[EuroSAT]{
     \centering
     \includegraphics[width=0.30\linewidth,height=0.22\linewidth]{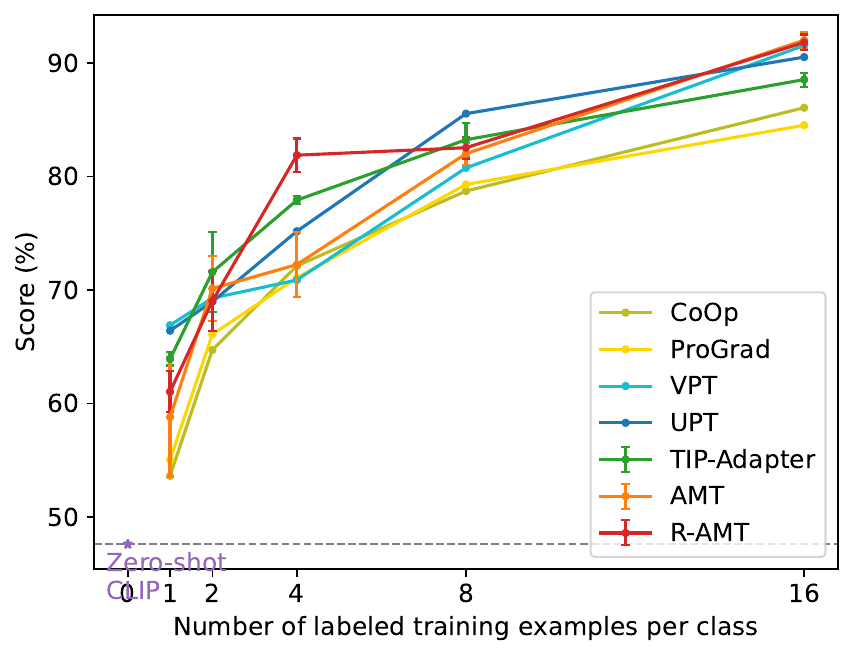}
     \label{fig:EuroSAT}
}
\hfill
\subfloat[Flowers102]{
     \centering
     \includegraphics[width=0.30\linewidth,height=0.22\linewidth]{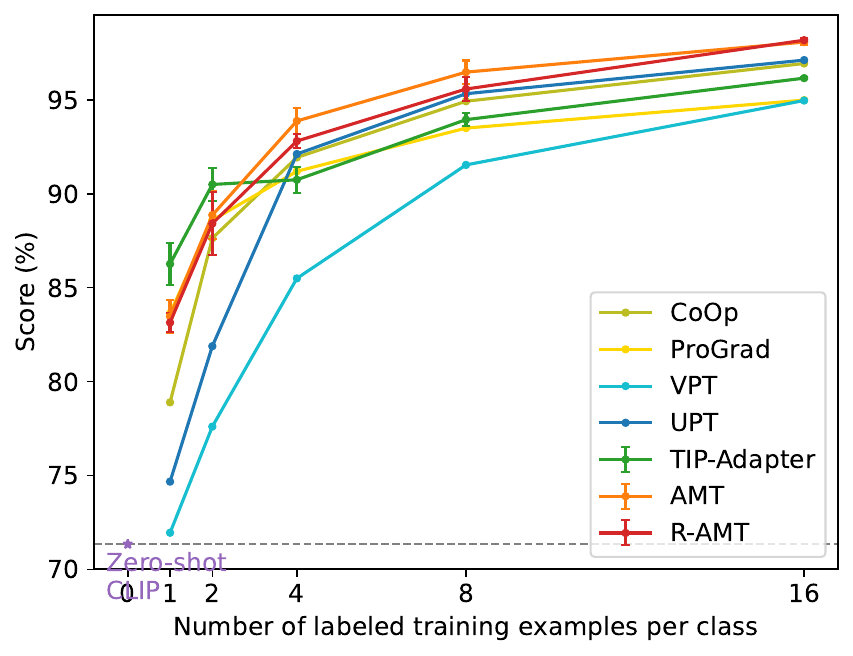}
     \label{fig:Flowers102}
}
\hfill
\subfloat[Food101]{
     \centering
     \includegraphics[width=0.30\linewidth,height=0.22\linewidth]{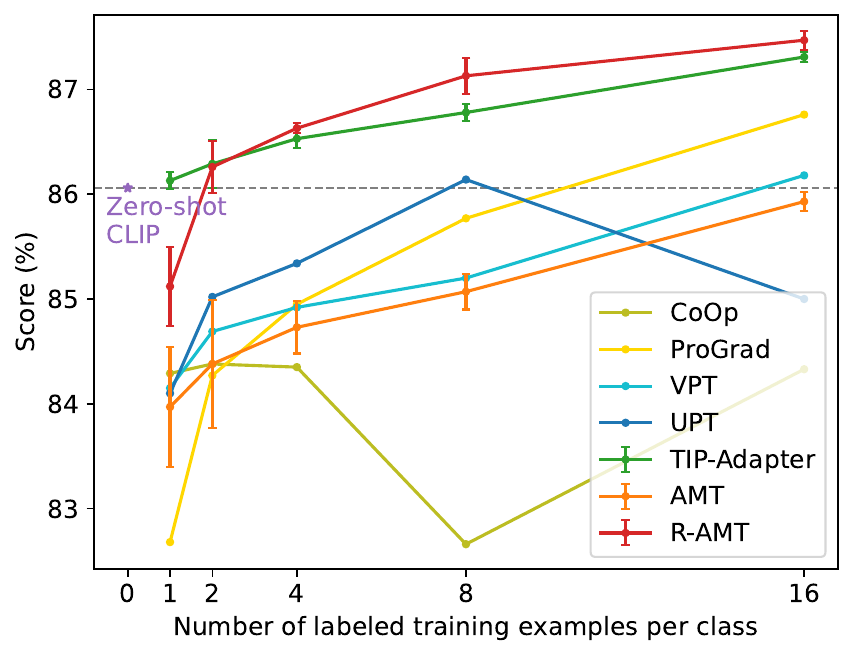}
     \label{fig:Food101}
}
\vspace{-10pt}
\subfloat[SUN397]{
     \centering
     \includegraphics[width=0.30\linewidth,height=0.22\linewidth]{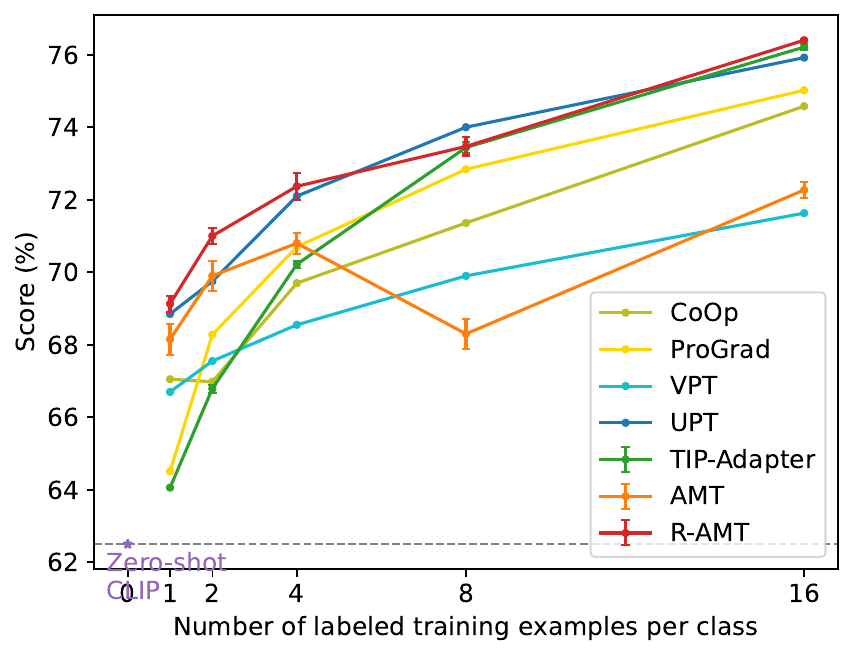}
     \label{fig:SUN397}
}
\hfill
\subfloat[OxfordPets]{
     \centering
     \includegraphics[width=0.30\linewidth,height=0.22\linewidth]{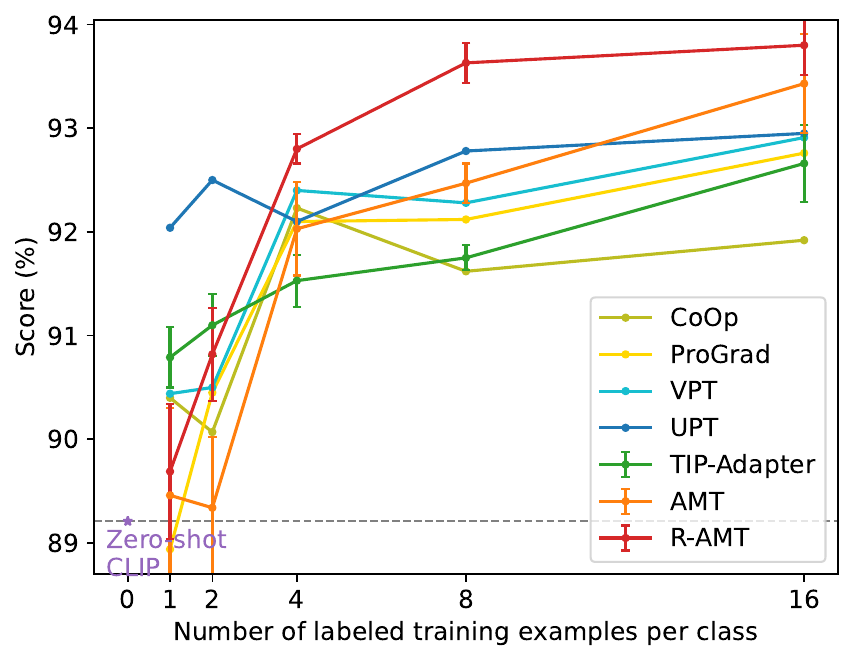}
     \label{fig:OxfordPets}
}
\hfill
\subfloat[DTD]{
     \centering
     \includegraphics[width=0.30\linewidth,height=0.22\linewidth]{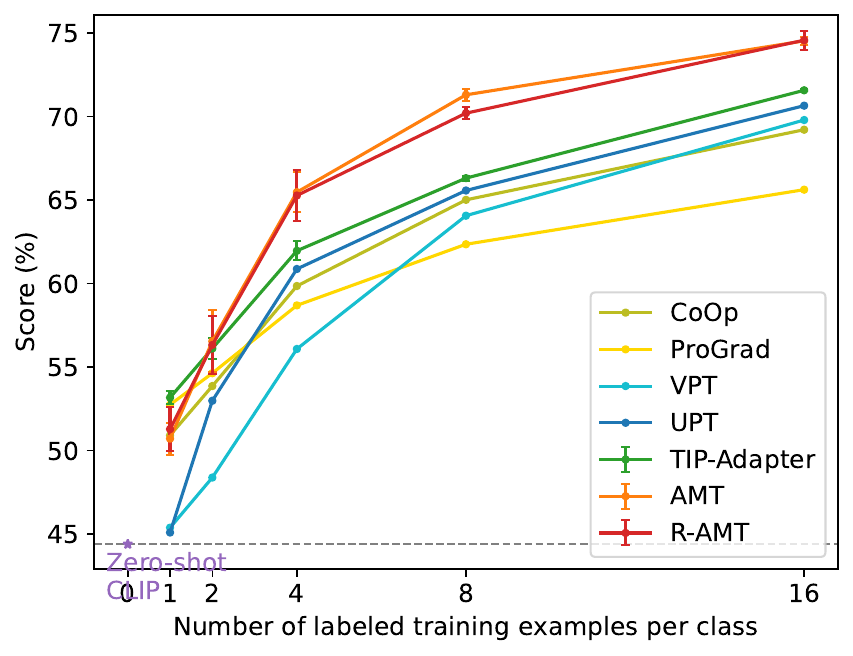}
     \label{fig:DTD}
}
\vspace{5pt}
\caption{ \textbf{Accuracy (\%) of few-shot learning, \ie, 16/8/4/2/1-shot, on the 11 datasets.} We report the average accuracy over three runs. For AMT, R-AMT, and TIP-Adapter, we demonstrate the \textit{error bar} in all figures.}
\label{fig:main_results_vit} 
\vspace{-.1in}
\end{figure*}

\noindent\textbf{Implementation Details}.
We transfer CLIP to the few-shot image classification task with AMT and R-AMT. Specifically, we use 1, 2, 4, 8, and 16-shot training sets to optimize the model and evaluate it on the full test set, following~\cite{radford2021learning}. For $n$-shot image classification, we random sample $n$ images per category for training. All results reported below are the average of three runs with different random seeds. All images are resized to $224\times 224$. Random cropping, resizing, and random horizontal flipping strategy are used for data augmentation.
We utilize ViT-B/16 as the visual backbone of CLIP. 
For a fair comparison, all experiments only use single text prompt, except learnable text prompt methods, \eg, for ImageNet and SUN397, the text prompt is set to be ``a photo of a [class].''
We adopt Adam optimizer for optimization.
The mask weights are initialized element-wise with $10^{-2}$. The threshold $\alpha$ is set to be $5\times10^{-3}$.
The $l$ in~\cref{eq.optimize_final} is set to 0.3 for datasets except for ImageNet, SUN397, and Food101 in 16-shot experiments. And $l$ is set to 1.0 in other experiments.

\subsection{Comparison to State-of-the-Art Methods}
\noindent
\textbf{Main Results on 11 Datasets.} We compare AMT and R-AMT with Zero-shot CLIP and five state-of-the-art methods on the 11 datasets as mentioned above, demonstrated in~\cref{fig:main_results_vit}. 
Zero-shot CLIP directly transfers to the downstream task without training.
The state-of-the-art methods include prompt tuning methods, \ie, CoOP~\cite{zhou2022learning}, VPT~\cite{jia2022visual}, UPT~\cite{zang2022unified}, ProGrad~\cite{zhu2022prompt}, and adapter tuning method TIP-Adapter~\cite{zhang2022tip}.
According to \cref{fig:Average}, the AMT and R-AMT outperform these methods on average over 11 datasets, which approves the ability of AMT and R-AMT to transfer CLIP to the downstream tasks. R-AMT achieves better performance compared with AMT. It indicates the gradient dropout regularity formalism is able to enhance the transfer ability of mask tuning in few-shot scenarios.


\noindent
\textbf{Results on base-to-new generalization setting.}
Following CoCoOP~\cite{zhou2022conditional}, we conduct experiments on base-to-new generalization setting. Concretely, the classes are split equally into the base and new classes on each dataset. The base classes are used for training. The $l$ is set to 1 in all base-to-new generalization experiments. The averaged results over 11 datasets are shown in~\cref{tab:Average_b2n}. The numerical experimental results on each dataset are shown in Supplementary.
Overall, R-AMT reaches the best performance, which surpasses the second best method CLIP-Adapter~\cite{gao2021clip} 2.00\% on the harmonic mean on average.
Notably, the AMT achieves quite high performance on the base classes. But the accuracy has significantly degraded (5.11\% on average) on the new classes compared with Zero-shot CLIP. 
We deem the degradation to be the result of overfitting since the amount of training data is too small for some datasets, \eg, Eurosat. 
The R-AMT achieves competitive results with AMT on base classes. However, the performance of R-AMT improves AMT by 3.04\% on average in new classes, which demonstrates the anti-overfitting ability of the proposed gradient dropout regularity formalism.

\begin{table}[t]
    \caption{\textbf{Comparison on the base-to-new generalization setting on the average over 11 datasets with 16 shots.} ``H'' denotes the harmonic mean of the accuracy on base and new classes. Thanks to gradient dropout regularity, R-AMT can efficiently maintain the knowledge of new classes while improving the anti-overfitting ability of the model to base classes. We report the average accuracy over three runs. The  error bar and performance of each dataset are provided in the supplementary materials.}
    \vspace{2pt}
    \centering\small
    \setlength{\tabcolsep}{13pt}
    \begin{tabular}{ccc|c}
    \toprule
        Method & Base & New & H \\
         \hline
       Zero-shot CLIP  &  69.34 & \textbf{74.22} & 71.70 \\
       CoCoOP~\cite{zhou2022conditional}  & 80.47 & 71.69 & 75.83 \\
       ProGrad~\cite{zhu2022prompt}  & 82.79 & 68.55 & 75.00 \\
       CLIP-adapter~\cite{gao2021clip}  & 82.62 & 70.97 & 76.35 \\
       \hline
       \multirow{1}{*}{AMT}  & \textbf{86.17} & 69.11 & \underline{76.70}  \\
       \multirow{1}{*}{R-AMT} & \underline{85.71} &\underline{72.15} & \textbf{78.35} \\
    \bottomrule
    \end{tabular}
    \label{tab:Average_b2n}
    \vspace{-8pt}
\end{table}

\noindent
\textbf{The robustness to distribution shift.}
We evaluate the out-of-distribution (OOD) ability of AMT and R-AMT by training them on ImageNet and evaluating on ImageNet-V2~\cite{recht2019imagenet} and Imagenet-Sketch~\cite{wang2019learning}, following~\cite{zhang2022tip}. The evaluating datasets have compatible categories with the training set. But the three datasets are different in semantics. The OOD experimental results are shown in~\cref{table:distribution_shift}. R-AMT achieves the best performance, which surpasses TIP-Adapter~\cite{zhang2022tip} 1.06\% on ImageNet-V2 and surpasses CoOP~\cite{zhou2022learning} 0.04\% on Imagenet-Sketch. This indicates the R-AMT is also capable of OOD tasks. Moreover, R-AMT boosts AMT 0.47\%, 0.94\%, and 0.91\% on ImageNet, ImageNet-V2, and Imagenet-Sketch, respectively. It further proves that R-AMT benefits from the gradient dropout regularity technique in terms of enhancing transfer and anti-overfitting ability.

\subsection{Combination with State-of-the-Art Methods}
\noindent
To prove the R-AMT is synergistic to existing parameter-efficient methods, we combine it with CoOP~\cite{zhou2022learning} and TIP-Adapter~\cite{zhang2022tip} on 11 datasets with 16 shots, as shown in~\cref{table:combine_sota_vit}. 
Concretely, we first load the binary masks trained with R-AMT  and multiply them with the original parameters of the image encoder of CLIP. Then we train the learnable contextual prompt or adapter following CoOP and TIP-Adapter.
Particularly, the few-shot training set for R-AMT, CoOP+R-AMT, and TIP-Adapter+R-AMT is the same.
For CoOP+R-AMT, the learned text prompt is randomly initialized and the length of the text prompt is set to 16, using the same training details as CoOP~\cite{zhou2022learning}. 
The CoOP+R-AMT boosts the performance of CoOP by 3.26\% on average.
This indicates the R-AMT provides a more reliable image encoder for learning better text prompts using CoOP.
In addition, this combination approach directly uses a mask that is optimized by hand-craft text and does not update this mask for the learnable text prompts from CoOP, resulting in not completely unleashing the potential of the mask for downstream tasks (\ie, not surpass the R-AMT).
For TIP-Adapter+R-AMT, the training details are also the same as the TIP-Adapter~\cite{zhang2022tip}.
The TIP-Adapter+R-AMT improves TIP-Adapter 3.13\% on average with 16 shots.
This verifies the ability of R-AMT to endow existing parameter-efficient methods with the ability to better adapt to the downstream task.

\begin{table}[t]
	\caption{\textbf{Comparison on robustness to distribution shift.}}
	\vspace{2pt}
	\centering\small
        \setlength{\tabcolsep}{2.2pt}
		\begin{tabular}{c|c|cc|c}
		\toprule
            \multirow{2}{*}{Method}  & Source & \multicolumn{2}{c|}{Target}  & \multirow{2}{*}{Average} \\
            & ImageNet & -V2 & -Sketch &  \\
            \hline
            
		Zero-shot CLIP & 66.73 & 60.83 & 46.15 & 57.90 \\
            Linear probe & 65.85 & 56.26 & 34.77 & 52.29 \\
            CoOP & 71.73 & 64.56 & 47.89 & 61.39 \\
            CLIP-adapter & 71.77 & 63.97 & 46.27 & 60.67 \\
            TIP-adapter & \textbf{73.08} &64.85 & 46.76 & \underline{61.56} \\
            \hline
            AMT & 72.60\std{0.12} & \underline{64.97}\std{0.11} & \underline{47.02}\std{0.13} & 61.53 \\
            R-AMT &\underline{73.07}\std{0.10} & \textbf{65.91}\std{0.34} & \textbf{47.93}\std{0.26} & \textbf{62.30}\\
            
		\bottomrule
		\end{tabular}
		\label{table:distribution_shift}
		
	\vspace{-8pt}
\end{table}

		
		
		

\begin{table*}[t]
	\caption{\textbf{Combining with the state-of-the-art methods on 16 shots.} Our mask tuning is synergistic with most existing parameter-efficient tuning methods (\eg, adapter tuning~\cite{zhang2022tip} and prompt tuning~\cite{zhou2022learning}) and can boost about $3\%$ performance on top of them.}
        \vspace{2pt}
	\centering\small
        \setlength{\tabcolsep}{3pt}
		\begin{tabular}{c|ccccccccccccc}
		\toprule
		\makebox[0.05\textwidth][c]{\rotatebox{45}{Method}} &
		\makebox[0.05\textwidth][c]{\rotatebox{45}{ImageNet}} &
		\makebox[0.05\textwidth][c]{\rotatebox{45}{Caltech101}} &
		\makebox[0.05\textwidth][c]{\rotatebox{45}{FGVCAircraft}} &
		\makebox[0.05\textwidth][c]{\rotatebox{45}{StanfordCars}} & 
            \makebox[0.05\textwidth][c]{\rotatebox{45}{Flowers102}} & 
            \makebox[0.05\textwidth][c]{\rotatebox{45}{OxfordPets}} &
		\makebox[0.05\textwidth][c]{\rotatebox{45}{Food101}} &
		\makebox[0.05\textwidth][c]{\rotatebox{45}{DTD}} &
		\makebox[0.05\textwidth][c]{\rotatebox{45}{EuroSAT}} &
		\makebox[0.05\textwidth][c]{\rotatebox{45}{UCF101}} &
		\makebox[0.05\textwidth][c]{\rotatebox{45}{SUN397}} & 
		\makebox[0.05\textwidth][c]{\rotatebox{45}{Average}}& 
		\makebox[0.05\textwidth][c]{\rotatebox{45}{Gain}}\\
 
		\hline 
		Zero-shot CLIP & 66.73 & 92.94 & 24.72 & 65.32 & 71.34 & 89.21 & 86.06 & 44.39 & 47.60 & 66.75 & 62.50 & 65.23 &- \\
		R-AMT & 73.07 & 97.00 & 58.47 & 85.93 & 98.17 & 93.80 & 87.47 & 74.57 & 91.80 & 86.93 & 76.40 & \textbf{83.96}&{\textcolor{red}{+18.73}}  \\
            \hline
		CoOP~\cite{zhou2022learning} & 72.01 & 95.47 & 43.29 & 82.91 & 96.93 & 91.92 & 84.33 & 69.21 & 86.05 & 82.25 & 74.58 & 79.90 &- \\
		CoOP+R-AMT& {73.35} & 96.70 & 56.37 & 85.63 & 97.83 & 93.20 & 86.13 & 73.03 & 90.20 & 86.87 & 75.45 &\textbf{83.16}&{~\textcolor{red}{+3.26}} \\ 
            \hline
		TIP-Adapter~\cite{zhang2022tip} & 73.08 & 95.63 & 45.20 & 83.04 & 96.15 & 92.66 & 87.31 & 71.57 & 88.53 & 84.24 & 76.21 & 81.24 &- \\
		TIP-Adapter+R-AMT &{74.28} & 96.97 & 61.07 & 86.27 & 97.80 & 94.07 & 87.43 & 74.77 & 91.50 & 86.93 & 76.97 & \textbf{84.37}&{~\textcolor{red}{+3.13}}\\
		
		\bottomrule
		\end{tabular}
		\label{table:combine_sota_vit}
		
	\vspace{-5pt}
	\end{table*}

\begin{table*}[t]
	\caption{\textbf{Effect of performing masking on different layers.} Attaching a binary mask to the multi-head self-attention layer (\ie, R-AMT) achieves the same performance as R-PMT but with lower computational effort.}
        \vspace{2pt}
	\centering\small
        \setlength{\tabcolsep}{4.8pt}
		\begin{tabular}{c|ccccccccccccc}
		\toprule
		\makebox[0.05\textwidth][c]{\rotatebox{45}{Method}} &
		\makebox[0.05\textwidth][c]{\rotatebox{45}{ImageNet}} &
		\makebox[0.05\textwidth][c]{\rotatebox{45}{Caltech101}} &
		\makebox[0.05\textwidth][c]{\rotatebox{45}{FGVCAircraft}} &
		\makebox[0.05\textwidth][c]{\rotatebox{45}{StanfordCars}} & 
            \makebox[0.05\textwidth][c]{\rotatebox{45}{Flowers102}} & 
            \makebox[0.05\textwidth][c]{\rotatebox{45}{OxfordPets}} &
		\makebox[0.05\textwidth][c]{\rotatebox{45}{Food101}} &
		\makebox[0.05\textwidth][c]{\rotatebox{45}{DTD}} &
		\makebox[0.05\textwidth][c]{\rotatebox{45}{EuroSAT}} &
		\makebox[0.05\textwidth][c]{\rotatebox{45}{UCF101}} &
		\makebox[0.05\textwidth][c]{\rotatebox{45}{SUN397}} & 
		\makebox[0.05\textwidth][c]{\rotatebox{45}{Average}}& 
		\makebox[0.05\textwidth][c]{\rotatebox{45}{Storage Space}}\\
 
		\hline 
		R-AMT & 73.07 & \textbf{97.00} & 58.47 & 85.93 & 98.17 & 93.80 & 87.47 & 74.57 & \textbf{91.80} & 86.93 & \textbf{76.40} & \textbf{83.96} & 6.7M \\
            R-MMT & \textbf{73.52} & 96.77 & 59.57 & \textbf{86.43} & 98.07 & \textbf{93.83} & 87.40 & 75.73 & 84.07 & 87.70 & 74.23 & 83.39 & 14M \\
            R-PMT & 73.48 & 96.63 & \textbf{60.30} & 86.33 & \textbf{98.27} & 93.77 & \textbf{87.50} & \textbf{75.60} & 88.20 & \textbf{87.33} & 76.12 & \textbf{83.96} & 19M \\
		
		\bottomrule
		\end{tabular}
		\label{table:different_layes_vit}
		
        \vspace{-.1in}
	\end{table*}

\subsection{Ablation Studies}
\label{sec:ab}
\noindent
\textbf{Analysis of Masking different layers.}
We conduct ablation studies on masking different layers of the image encoder.~\cref{table:different_layes_vit} shows the results on 16 shots over 11 datasets. Concretely, we apply the binary masks on all weight matrices of convolutional layers and fully connected layers when training R-PMT. For R-AMT, the binary masks are applied on the multi-head self-attention (MHSA) layers, while for R-MMT, the binary masks are applied on the multilayer perceptron (MLP) layers.
R-AMT achieves equal performance with R-PMT on the average of 11 datasets, which surpasses R-MMT 0.57\%. 
But the R-AMT only uses 6.7M for storing the trained model, which is 12.3M less than R-PMT.
Moreover, we find the R-AMT achieves superior performance when there are limited training classes, \eg, EuroSAT.
Thus, we deem the R-AMT to be a more practical method.

\noindent
\textbf{Influence of gradient dropout regularity.}
We explore the influence of gradient dropout regularity with 16 shots ImageNet. The experimental results are shown in~\cref{table:graddrop}.
The proposed gradient dropout regularity requires the guidance of KL divergence. Thus, we conduct an ablation study by directly adding the KL loss $\mathcal{L}_{kl}$ with the cross-entropy loss $\mathcal{L}_{ce}$ to training the binary mask, which is termed as AMT+KL loss. 
It shows that if we directly add these two losses, the accuracy drops by 0.68\% on 16-shot ImageNet. Because the $\mathcal{L}_{ce}$ aims to transfer the model to downstream tasks, while the $\mathcal{L}_{kl}$ requires the disparity between the classification logits of AMT and CLIP is not large.
Directly adding $\mathcal{L}_{kl}$ with $\mathcal{L}_{ce}$ limits the transfer ability of AMT.
AgreeGrad~\cite{mansilla2021domain} adopts gradient surgery to solve the domain conflict, but it excessively believes in previous knowledge from KL loss, resulting in performance degradation.
Recently, ProGrad~\cite{zhu2022prompt} proposes a gradient projection method for training text prompts. This gradient projection method and our gradient dropout regularity technique both require the guidance of KL divergence.
Thus, we employ the gradient projection method to train our AMT for comparison, named AMT+ProGrad. AMT+ProGrad surpasses AMT by 0.10\%, but is 0.37\% lower than R-AMT ($l$=1.0).
It indicates the gradient projection technique can help mask tuning. But the improvement is limited since all conflict gradients are forced to be projected in the vertical direction.
The gradient dropout regularity adds some level of randomness to the gradient guided by the KL divergence, which helps the model generalize better to downstream tasks.
Moreover, we analyze the influence of $l$ in the gradient dropout regularity technique. 
A smaller $l$ implies a higher probability of CE-related gradient maintenance, which divers the binary masks more sparse.
The R-AMT achieves the best performance on 16-shot ImageNet when $l=1.0$. When $l$ is small than 1.0, The performance degradation is caused by the leak through gradients of $\mathcal{L}_{ce}$, which conflicts with the general knowledge of CLIP.

 \begin{table}[t]
	\caption{\textbf{Ablation studies on gradient dropout regularity strategy.} The proposed gradient dropout regularity can make better use of general knowledge of KL loss while exploring the knowledge of downstream data. ``$l$'' controls the level of agreement in CE Loss.
 }
        \vspace{2pt}
	\centering\small
        \setlength{\tabcolsep}{5pt}
		\begin{tabular}{c|c|c|c|c}
		\toprule

            Method & $l$ & Accuracy & Gain& Sparsity\\
            \hline 
            Zero-shot CLIP & - & 66.73 & - & -  \\
            \hline 
            AMT         & - & 72.60\std{0.12} & - & 2.64  \\
            AMT+KL loss & - & 71.92\std{0.06} & \textcolor{green}{-0.68} & 2.58 \\
            AMT+AgreeGrad~\cite{mansilla2021domain} & - & 68.82\std{0.09} & \textcolor{green}{-3.78} & 1.73 \\
            AMT+ProGrad~\cite{zhu2022prompt} & - & 72.70\std{0.22} & \textcolor{red}{+0.10} & 2.67  \\
            \hline
            R-AMT & 1.0 & \textbf{73.07}\std{0.10} & \textcolor{red}{+0.47}& 2.45 \\
            R-AMT & 0.8 & 72.97\std{0.13} & \textcolor{red}{+0.37}& 2.50 \\
            R-AMT & 0.5 & 72.95\std{0.19} & \textcolor{red}{+0.35} & 2.56 \\
            R-AMT & 0.3 & 72.87\std{0.05} & \textcolor{red}{+0.27} & 2.59 \\
            R-AMT & 0.1 & 72.67\std{0.12} & \textcolor{red}{+0.07} & 2.61 \\

		\bottomrule
		\end{tabular}
		\label{table:graddrop}
  \vspace{-10pt}
  \end{table}

\noindent
\textbf{Analysis of hard threshold $\alpha$.} We conduct ablation study on the hard threshold $\alpha$ with the initial value of mask weights fixed in~\cref{table:ablation_alpha}. It shows the binary masks are sparser as $\alpha$ gets larger. The R-AMT achieves the best accuracy when $\alpha=5\times10^{-3}$. We deem that some redundant information still has not been masked when $\alpha=4\times10^{-3}$. Thus, this information still influences the performance of the model in the downstream task. When $\alpha=6\times10^{-3}$, some valuable parameters are moved out by the binary masks. It caused performance degradation.
	
 \begin{table}[t]
	\caption{\textbf{Effect of hard threshold $\alpha$ on 16-shot ImageNet}. The threshold determines the sparsity of model.
 }
        \vspace{2pt}
	\centering\small
        \setlength{\tabcolsep}{10pt}
		\begin{tabular}{c|c|c|c}
		\toprule

            $\alpha$  & $4\times10^{-3}$ &  $5\times10^{-3}$ & $6\times10^{-3}$ \\
            \hline 
            Accuracy      & 72.87\std{0.06}  & \textbf{73.07}\std{0.10}& 72.91\std{0.14}  \\
            Sparsity  & 1.99 & 2.45 & 3.12\\
		\bottomrule
		\end{tabular}
		\label{table:ablation_alpha}
\vspace{-8pt}
  \end{table}


%% file: sections/5.conclusion.tex
\section{Conclusion}\label{sec:conclusion}
In this work, we design a new type of tuning method, termed regularized mask tuning, that masks the network parameters under a learnable selection. 
Specifically, we first identify a set of parameters that are key to a given downstream task, then attach a binary mask to this parameter, and finally optimize these masks on the downstream data with the parameters frozen.
When updating the mask, we introduce a novel gradient dropout strategy to regularize the parameter selection, to prevent the model from forgetting and overfitting.
Extensive experiments demonstrate that our method consistently outperforms existing methods and is synergistic with them.
Future work will explore applying mask tuning to other visual tasks such as segmentation.

%% file: sections/6.ref.tex
{\small
\bibliographystyle{ieee_fullname}
\bibliography{ref}
}

%% file: sections/appendix.tex
\newcommand{\suppstd}[1]{$\pm\text{#1}$}
\clearpage
\appendix
\onecolumn

%

%
\noindent{\large Supplementary Materials Organization:}

\noindent\DoToC 


\section{Limitations and Broader Impact}
\noindent\textbf{Broader Impact.} As for positive impact,  we design a novel mask-tuning method strategy to select a subset of network parameters in a pre-trained model for few-shot visual recognition tasks. 
The learned mask-tuning method can further boost the transfer capacity of the existing prompt-based and adapter-based methods.

\noindent\textbf{Limitations.} As for limitations, our method, as a general method, has not been verified on open-world detection and segmentation tasks due to limited computational resources. We leave this exploration in the future.

\section{Preliminaries of CLIP-related Tuning Methods}
CLIP~\cite{radford2021learning} mainly consists of two components: text encoder $\boldsymbol{G}_T$ and image encoder $\boldsymbol{G}_I$, which are designed to project image and text into the same feature embedding space. Concretely, the text encoder is built with the transformer for extracting text features. Meanwhile, the image encoder is used to extract image features that have the same channel dimension as the text features.
The architecture of the image encoder can be ResNet \cite{he2016deep} or ViT \cite{dosovitskiy2020image}. Cosine similarity between text and image features is utilized for alignment in CLIP.
The CLIP, benefiting from the 400 million text-image pairs from the web and the multi-modality structure, achieves exceptional zero-shot transfer capacity in the downstream tasks. 

To improve the transferability in the various downstream tasks, some parameter-efficient studies based on these V\&L models, \textit{e.g.}, adapter~\cite{gao2021clip,zhang2022tip} or prompt~\cite{zhou2022learning,zang2022unified}, are proposed. Specifically, Zhou \etal~\cite{zhou2022learning} change the hand-craft text prompt to a task-specific learnable \textit{text prompt}, which can be formulated as ``$[\boldsymbol{T}_1] [\boldsymbol{T}_2] \cdots [\boldsymbol{T}_l][class]$''. Here, the $l$ refers to the length of the learnable text prompt. The text encoder extracts text features $\theta\prime_i$ from the learned text prompt to match the image features, the same as~\cref{eq.possibility}.
The learnable text prompt is optimized with cross-entropy classification loss. 
Zang \etal \cite{zang2022unified} introduce a unified prompt into the text and image branches. The unified prompt is also a set of learnable parameters $\boldsymbol{U}\in \mathcal{R}^{d*l}$, where the $d$, $l$ denote the dimension and length of the prompt, respectively. Then the unified prompt is refined by a transformer layer and split into two parts to complete the image and text input, which can be formulated as follows:
\begin{gather}
    \label{eq.unified_prompt}
    \boldsymbol{U}' = \text{transformer}(\boldsymbol{U}),  \\
    \{\boldsymbol{U}_t, \boldsymbol{U}_v\} = \boldsymbol{U}',
\end{gather}
where the $\boldsymbol{U}_t, \boldsymbol{U}_v$ denote text prompt and visual prompt, respectively. Then the prompts are combined with text or image to be used as input for CLIP.
For the adapter-based method, Zhang \etal~\cite{zhang2022tip} build an adapter following the image encoder of CLIP. Given $S$-shot training data, the weights of adapter $\boldsymbol{A}$ are initialized with the few-shot image features $\boldsymbol{F}_I\in\mathcal{R}^{m\times d}$ encoded by the image encoder $\boldsymbol{G}_I$. The ground truth labels of images are converted into a one-hot vector $\boldsymbol{L}_I\in\mathcal{R}^{m\times k}$. The possibility of assigning image $\vx_j$ to class $\vy_i$ can be formulated as:
\begin{gather}
    \label{eq.tip}
    p_t(\vy=i\mid \vx_j)=\alpha \boldsymbol{A}(\boldsymbol{f}_j)\boldsymbol{L}_I^i+p(\vy=i \mid \vx_j), \\
    \boldsymbol{A}(\boldsymbol{f}_j) = exp(-\beta(1-\boldsymbol{f}_j\boldsymbol{F}^\mathrm{T}_I)),
\end{gather}
where the $\alpha$ and $\beta$ are hyper-parameters, $\boldsymbol{L}_I^i\in\mathcal{R}^{m\times 1}$ denotes $i$-th column of $\boldsymbol{L}_I$, which corresponds to class $i$. The TIP-Adapter performs better when fine-tuning the adapter $\boldsymbol{A}$ with $S$-shot training. Our method is orthogonal to the most existing parameter-efficient adaption methods (\eg, adapter and prompt) and endows them the ability to customization on downstream needs.

\section{Method Details}

Different from the common tuning methods that adopt the image/text prompts or adapter modules, we design a new type of tuning method, termed mask tuning method, which masks the network parameters under a learnable selection. Specifically, we apply binary masks on CLIP to search a subset of pre-trained parameters relevant to downstream tasks. 
In this way, to better understand the proposed mask tuning method, let's take a fully connected layer as an example (other layers, such as convolution and attention, are the same operation). Specifically, given a fully connected layer, the input and output of which are $\vx_{\vtheta_i}\in \mathcal{R}^{c_{in}}$ and $\vy_{\vtheta_i}\in \mathcal{R}^{c_{out}}$, respectively. The $c_{in}$ denotes the channel dimension of input, and the $c_{out}$ refers to the channel dimension of output. The weight matrix of the fully connected layer is $\vtheta_i= \mathcal{R}^{c_{out}\times c_{in}} $, which can be expanded as:
\begin{align}
    \vtheta_i & = \left[\begin{array}{ccc}
{\theta_{1,1}} & \cdots & {\theta_{1, c_{in}}} \\
\vdots & \ddots & \vdots \\
{\theta_{c_{out}, 1}} & \cdots & {\theta_{c_{out}, c_{in}}}
\end{array}\right].
\end{align}
The fully connected layer can be formulated as follows:
\begin{equation}
    \label{eq.linear}
    \vy_{\vtheta_i} = \vtheta_i \cdot\vx_{\vtheta_i}+\boldsymbol{b},
\end{equation}
where $\vb \in \mathcal{R}^{c_{out}} $ is the bias vector. For each weight matrix, we employ a learnable matrix $\mM$ with initializing value $\pi$, which has the same shape as the weight matrix $\theta$. We set a hard threshold $\alpha$ to binarized the learnable matrix $\mM$ as follows:
\begin{equation}
    \label{eq.mask}
    {m}^{bin}_{i,j} = \begin{cases}
    1, & \text{if  } {m}_{i,j}\geq \alpha \\
    0, & \text{if  } {m}_{i,j}< \alpha
    \end{cases}
    ,
\end{equation}
where the ${m}_{i,j}$ denotes the parameter in the $i$-th row and $j$-th column of learnable matrix $\mM$, and $m^{bin}_{i,j}$ denotes the corresponding parameter in binary mask $m^{bin}$. Then the updated weight matrix $\theta'$ is obtained as following:
\begin{align}
    \vtheta_{i, \gM} := \vtheta_i\odot \mM^{bin}, & 
\end{align}
where the $\odot$ refers to Hadamard product.

Since previous works \cite{zhao2020masking,mallya2018packnet} have shown that training task-specific bias has not a significant improvement for the downstream tasks, we only apply the binary mask on the weight matrix to lower the computation cost. Thus, the updated fully connected layer can be formulated as $\vy_{\vtheta_i} = \vtheta_{i,\mM} \cdot\vx_{\vtheta_i}+\boldsymbol{b}$. This method can be easily extended to convolutional layers, where we also only apply binary masks on the weight matrix.
The binary mask is optimized with the cross-entropy classification loss.
Importantly, since the binarized function shown in \cref{eq.mask} is non-differentiable, we use the gradient of $m^b$ as a noisy estimator to update the learnable matrix $m$, following the previous work \cite{zhao2020masking,lin2020dynamic}. The optimization can be formulated as:
\begin{equation}
    \label{eq:m_grad_pmt}
    m_{i,j} \gets  m_{i,j}-\gamma \frac{\partial \mathcal{L}_{ce}}{\partial m^{bin}_{i,j}},
\end{equation}
where the $\gamma$ denotes the learning rate that controls the sparsity of mask, and the $ \mathcal{L}_{ce}$ denotes the loss value obtained from the Cross-Entropy (CE) loss function.
Then, we analyze the change in mask weight $\mM$ for each layer after training on the target dataset with the CE loss, \ie, $\Delta = \sum \gamma* \frac{\partial\ce}{\partial\mM}$. Multi-head self-attention (MSHA) layers play an important role in the mask-tuning method. This suggests binary attention mask $\boldsymbol{M}^{bin}_a$ plays a significant role during fine-tuning, and we leverage that in our method.
We mathematically rewrite the~\cref{eq:m_grad_pmt} formulated as:
\begin{equation}
    \label{eq:m_grad}
    m_{a;i,j} \gets  m_{a;i,j}-\gamma \frac{\partial \mathcal{L}_{ce}}{\partial m^{bin}_{a;i,j}},
\end{equation}
where $m_{a;i,j}$ represents the mask value for the index ${i,j}$ of the mask matrix $\mM_a$ in the MHSA layer.
Then, we calculate the Gradient Retaining Purity $\mathcal{P}$ as follows 
\begin{equation} 
\label{eq:psp_appendix} 
\mathcal{P} = \frac{1}{2}\left(1+\frac{\text{sgn}(\gce)\left(\gce + \gkl\right)}{|\gce| + |\gkl|}\right).
\end{equation}
The optimization can be formulated as:
\begin{equation}
    \label{eq.optimize_final_appendix}
    m_{a;i,j} \gets  m_{a;i,j}-\gamma * (1-l+l*\mathcal{I}[\mathcal{P}>U]) * \frac{\partial \mathcal{L}_{\text{ce}}}{\partial m_{a;i,j}^{bin}},
\end{equation}
where $U$ is a tensor composed of i.i.d $U(0,1)$ random variables and $l \in [0,1]$ is a leak parameter. $l<1$ means we allow $\gce$ leak through.



\section{Experimental Details}

\subsection{Statistic of Datasets}
We conduct experiments on 11 publicly available image classification datasets following CoOP~\cite{zhou2022learning}. The datasets including ImageNet~\cite{deng2009imagenet}, FGVCAircraft~\cite{maji2013fine}, StanfordCars~\cite{Krause_2013_ICCV_Workshops}, Flowers102~\cite{nilsback2008automated}, Caltech101~\cite{fei2004learning}, DTD~\cite{cimpoi2014describing}, EuroSAT~\cite{helber2019eurosat}, Food101~\cite{bossard2014food}, UCF101~\cite{soomro2012ucf101}, OxfordPets~\cite{parkhi2012cats}, and SUN397~\cite{xiao2010sun}. 
For distribution shift experiments, we use ImageNet as the source dataset, while ImageNetV2~\cite{recht2019imagenet} and  ImageNet-Sketch~\cite{wang2019learning} are used as the target dataset. 
We report the detailed statistics of the 13 datasets in \cref{table:dataset_statistic}.

\begin{table*}[htbp]
    \small
    \centering
    \caption{The detailed statistics of datasets used in experiments. }
    \label{tab:dataset}
    \resizebox{0.8\textwidth}{!}{
    \begin{tabular}{lcccc}
    \toprule
Dataset                  & Classes  & Training size  & Testing size & Task \\ \hline
Caltech101~\cite{fei2004learning} & 100 & 4,128 & 2,465& Object recognition \\
DTD~\cite{cimpoi2014describing}& 47 & 2,820 & 1,692 &  Texture recognition\\ 
EuroSAT~\cite{helber2019eurosat}& 10 & 13,500 & 8,100 & Satellite image recognition \\ FGVCAircraft~\cite{maji2013fine} & 100 & 3,334 & 3,333 & Fine-grained aircraft recognition\\
Flowers102~\cite{nilsback2008automated} & 102 & 4,093 & 2,463 & Fine-grained flowers recognition \\ Food101~\cite{bossard2014food} & 101 & 50,500& 30,300 & Fine-grained food recognition  \\ ImageNet~\cite{deng2009imagenet} & 1,000 & 1.28M & 50,000 & Object recognition \\ OxfordPets~\cite{parkhi2012cats} & 37  & 2,944 & 3,669 & Fine-grained pets recognition \\ StanfordCars~\cite{Krause_2013_ICCV_Workshops} & 196 & 6,509 & 8,041 & Fine-grained car recognition \\
SUN397~\cite{xiao2010sun}& 397& 15,880 & 19,850 & Scene recognition\\ 
UCF101~\cite{soomro2012ucf101}& 101 & 7,639 & 3,783 & Action recognition\\
\hline
ImageNetV2~\cite{recht2019imagenet} & 1,000 & - & 10,000 & Robustness of collocation  \\
ImageNet-Sketch~\cite{wang2019learning} & 1,000 & - &50,889 & Robustness of sketch domain\\
    \bottomrule
    \end{tabular}
    }
    \label{table:dataset_statistic}
\end{table*}

		
		

\subsection{More Implementation Details }\label{app: imple}
We use single hand-craft prompt as text input when applying mask tuning, following~\cite{radford2021learning}. Specifically, for ImageNet and SUN397, the text prompt is set to be ``a photo of a [class].''. For fine-grained classification datasets, a task-relevant sentence is added, \eg, the text prompt is ``a photo of a [class], a type of flower.'' for Flowers102 dataset. For other datasets, the text prompt is set to be a task-related context, \eg, for UCF101, the text prompt is ``a photo of a person doing [class].''
We adopt Adam optimizer with CosineAnnealingLR schedule for optimization.
For ImageNet, the maximum epoch is set to 10, the learning rate is set to 3e-5.
For other datasets, the maximum epoch is set to 30, and the learning rate is set to 8e-5.
The few-shot classification task provides limited training data for fine-tuning model, which may lead to overfitting. Thus, we fix $l$ in~\cref{eq.optimize_final} to be 1 for 8/4/2/1-shot experiments to enhance the anti-overfitting ability of R-AMT.
For 16-shot classification task, we observe AMT surpasses Zero-shot CLIP by 18.23\% on average across 11 datasets. The upstream information introduced by KL loss may limit the transfer ability of our method. So we set $l=0.3$ for 16-shot experiments to allow the gradients from CE loss leak through.
However, considering the large amount of testing data may result in relatively large distribution gap between testing and few-shot training data. We fix $l=1$ for the ImageNet, SUN397, and Food101 datasets in 16-shot experiments.
The code of our method is based on CoOP~\cite{zhou2022learning}. We conduct experiments on 1 NVIDIA A100 GPU. All reported results are the average of three runs with different seeds. Moreover, since the learned binary masks by R-AMT are constructed by binary values. We treat each binary element as a bit and encode every 8 bits to a byte for storage, which greatly saves the storage space of the binary masks.

\section{More Experimental Analysis}
In this section, we report the average accuracy over three runs and demonstrate the error bar in figures and tables. ``error bar'' refers to standard deviation.
\subsection{Computation Cost Evaluation }\label{app:cost}

As shown in~\cref{tab:time}, we provide the comparison of the training time and inference time of existing SOTA methods (\eg, CoOp, CoCoOP, Tip-Adapter), AMT, and our R-AMT. We report the one-epoch time training on the 16-shot setting of the ImageNet dataset and the number of images processed by the model in 1 second (\ie, Frames Per Second (FPS)). Compared witn the Tip-adapter, AMT reduces the $10.3$ FPS inference speed and requires an extra $25.0$ FPS training time, which is acceptable given the performance improvement.
\begin{table*}[htbp]
\centering
    \caption{The training and inference time comparison.}
    \label{tab:time}
	 \setlength{\tabcolsep}{4pt}
	\begin{tabular}{l|ccccc}
	 \toprule
Settings                    & CoOp~\cite{zhou2022learning}& CoCoOP~\cite{zhou2022conditional} & Tip-Adapter~\cite{zhang2022tip} & AMT   & R-AMT \\ \hline
Training Time (images/s)           &  7.14   &    11.11 &     \textbf{50.00} &      25.00    &     15.87  \\
Inference Time (images/s)   & 7.45 & 12.21  & 51.81       & \textbf{62.11} & \textbf{62.11} \\
    \bottomrule
    \end{tabular}
\end{table*}

\subsection{Sparsity and Performance Comparison with Zero-shot CLIP}
\begin{figure}[htbp]
    \centering
    \includegraphics[scale=0.4]{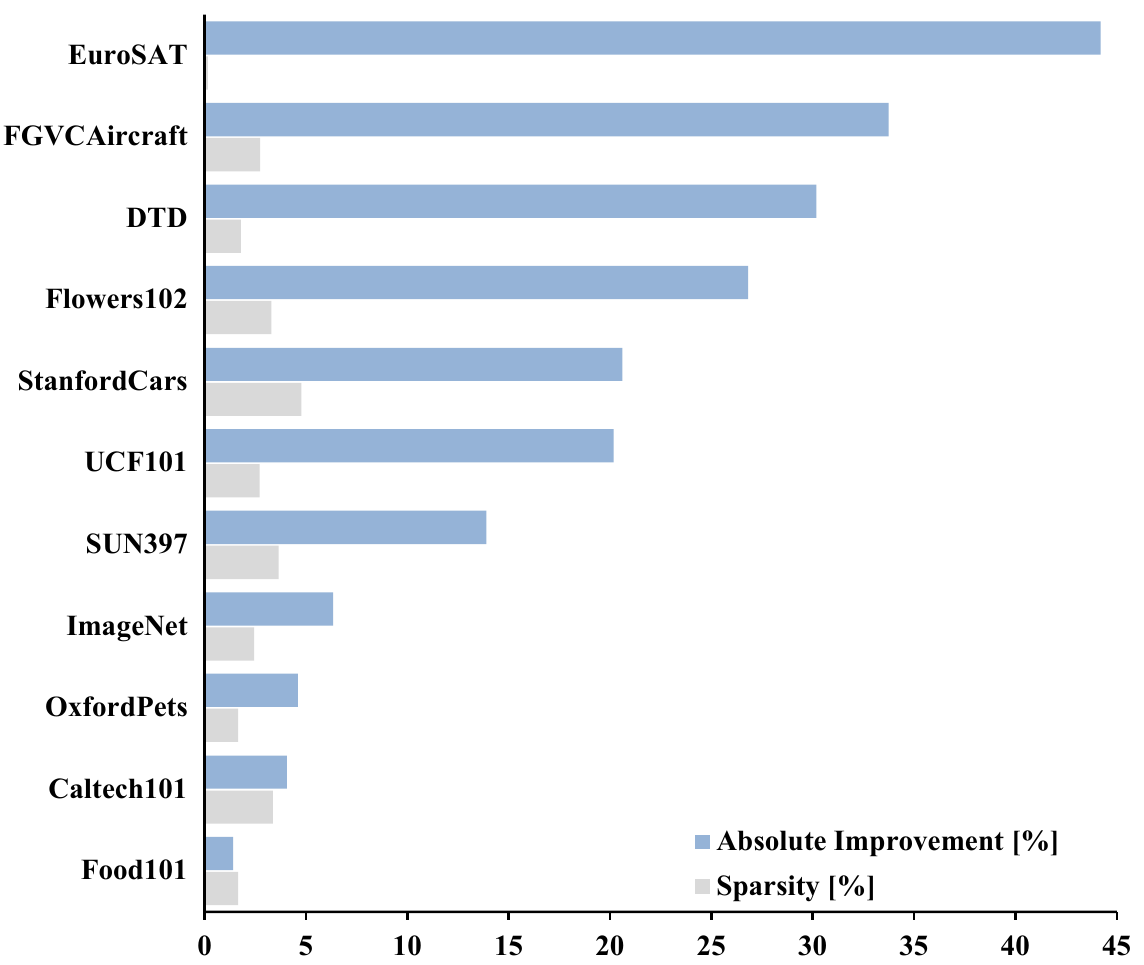}
    \caption{Comparison with Zero-shot CLIP in terms of accuracy and sparsity on 16-shot datasets. The sparsity means the percentage of the number of discarded parameters (mask=0).
    }
    \label{fig:compare_zs_vit}
    \vspace{-.1in}
\end{figure}

In~\cref{fig:compare_zs_vit}, we demonstrate the absolute improvement of R-AMT compared with Zero-shot CLIP and the sparsity of binary masks on the 16-shot setting. The R-AMT boosts the performance of Zero-shot CLIP on all datasets. Significant improvements are achieved on the EuroSAT and FGVCAircraft dataset, which reach 44.20\% and 33.75\%, respectively. 
The most sparse binary mask is obtained on StanfordCars. After setting 4.77\% parameters to 0, the R-AMT improves Zero-shot CLIP 20.61\% on accuracy.
In total, we manage to deliver \textbf{18.73\%} performance improvement compared to the zero-shot CLIP via masking an average of only \textbf{2.56\%} parameters.
It proves that the pre-trained weights contain some unnecessary information for the downstream task, which may harm the transfer ability of the pre-trained model. 

\subsection{Text Prompt Ensembling}\label{sec:prompt_ensembling}
We utilize the prompt ensembling of 7 templates to construct the text input on ImageNet, following TIP-Adapter~\cite{zhang2022tip}.
In~\cref{table:prompt_esambling}, we report the accuracy of R-AMT and R-PMT on 16-shot ImageNet. The R-AMT and R-PMT boost Zero-shot CLIP 4.76\% and 5.09\% in terms of the accuracy, respectively. The R-PMT improves TIP-Adapter by 0.13\% performance. It further indicates the effectiveness of mask tuning in fine-tuning CLIP.
Moreover, we combine R-AMT and R-PMT with TIP-Adapter on 16-shot ImageNet. The R-AMT+TIP-Adapter and R-PMT+TIP-Adapter both surpass TIP-Adapter. It means the image encoder assembled a learned binary mask extracts more distinctive image features in the downstream classification task.

\begin{table}[htbp]
\caption{Classification accuracy (\%) on 16-shot ImageNet when using prompt ensembling of 7 templates for text prompt.}
\centering
\resizebox{0.7\linewidth}{!}{
\begin{tabular}{c|c|c|c}
\toprule
Methods & \multirow{1}{*}{Zero-shot CLIP} & \multirow{1}{*}{R-AMT} & \multirow{1}{*}{R-PMT}\\
\hline
Accuracy (\%) & 68.73 & 73.49~\textcolor{red}{(+4.76)} &  73.82~\textcolor{red}{(+5.09)} \\ 
Error Bar & - & \suppstd{0.10} &  \suppstd{0.20} \\ 
\toprule
Methods &\multirow{1}{*}{TIP-Adapter~\cite{zhang2022tip}}& \multirow{1}{*}{R-AMT + TIP-Adapter~\cite{zhang2022tip}}& \multirow{1}{*}{R-PMT + TIP-Adapter~\cite{zhang2022tip}}\\ 
\hline
Accuracy (\%) & 73.69 & 74.20~\textcolor{red}{(+0.51)}  & 74.22~\textcolor{red}{(+0.53)} \\
Error Bar & - & \suppstd{0.22} &  \suppstd{0.53} \\ 
\bottomrule
\end{tabular}
}    
\vspace{-.1in}
\label{table:prompt_esambling}

\end{table}

\subsection{Different Vision Backbones} 
In~\cref{table:vision_backbone}, we report the results of implementing R-AMT and R-PMT with different vision backbones of CLIP on 16-shot ImageNet, including ResNet50, ResNet101, ViT-B-16, and ViT-B-32. Concretely, for R-PMT, we apply the binary masks on all weight matrices of convolutional layers and fully connected layers.
We observe that R-PMT achieves the best accuracy on all kinds of vision backbones on 16 shots ImageNet.
When utilizing ResNet50, ResNet101, ViT-B-16, and ViT-B-32 as vision backbones, the R-PMT outperforms the second-best method by 0.68\%, 1.13\%, 0.40\%, and 0.88\%, respectively. 
These results demonstrate that the binary tuning is superior to the prompts tuning and adapter tuning.
When using ViT as the visual backbone, R-AMT achieves competitive results with R-PMT but introduces fewer parameters. Thus, we still recommend R-AMT when ViT is the visual backbone of CLIP. 

\begin{table}[htbp]
	\caption{Comparison with the state-of-the-art methods with different vision backbones on 16-shot ImageNet.}
	
	\centering
	\resizebox{0.65\linewidth}{!}{
		\begin{tabular}{c|c|c|c|c}
		\toprule
		\multirow{1}{*}{Method} & \multirow{1}{*}{ResNet50} &
		\multirow{1}{*}{ResNet101}  &
		\multirow{1}{*}{ViT-B/16}&
		\multirow{1}{*}{ViT-B/32}\\ 
		\hline 
		
		Zero-shot CLIP & 58.18  & 61.62   & 66.73 &  62.05 \\
		VPT~\cite{jia2022visual} & - & - & 70.57 &- \\
		CoOP~\cite{zhou2022learning} & 62.90 & \underline{66.60} & 71.92& 66.85  \\
		CLIP-Adapter~\cite{gao2021clip} & 63.59  & 65.39  & 71.13 & 66.19  \\
		TIP-Adapter~\cite{zhang2022tip} & \underline{64.17} & 66.42  & \underline{73.08} &  \underline{67.12} \\
		UPT~\cite{zang2022unified} & - & - & 72.63 &- \\
            PLOT~\cite{chen2022prompt} & 63.01 & - & - & -\\
		\hline
            R-AMT & - & - & 73.07 & 67.84 \\
		 R-PMT & \textbf{64.85~\textcolor{red}{(+0.68)}} & \textbf{67.73~\textcolor{red}{(+1.13)}}  & \textbf{73.48~\textcolor{red}{(+0.40)}} & \textbf{68.00~\textcolor{red}{(+0.88)}}  \\
		\bottomrule
		\end{tabular}
		}  
		\label{table:vision_backbone}
		\vspace{-.1in}
\end{table}

\subsection{Analyzing the Differences between Fine-tuning the Entire Network and Tuning the Mask}
In~\cref{table:compare_ft}, we report the performance of fine-tuning and mask tuning the image encoder of CLIP on 16-shot ImageNet. The ``Fine-tuning'' denotes fine-tuning the whole image encoder.
We observe fine-tuning the entire network results in performance degradation compared to Zero-shot CLIP. Tuning the Mask also demonstrates clear advantages over the linear probe model. It is also clear that the gaps in the extremely low-data regime between fine-tuning the entire network and tuning the mask, suggest that mask tuning is much more effective than learning a linear classifier from scratch or fine-tuning the entire network for few-shot learning.

\begin{table}[htbp]
\caption{Comparison with Fine-tuning on 16-shot ImageNet.}
\label{table:compare_ft}
\centering
\resizebox{0.65\linewidth}{!}{
\begin{tabular}{c|c|c|c|c|c}
\toprule
Methods & \multirow{1}{*}{Zero-shot CLIP} & \multirow{1}{*}{Fine-Tuning} & \multirow{1}{*}{Linear Probe} & \multirow{1}{*}{AMT} & \multirow{1}{*}{R-AMT}\\
\hline
Accuracy & 66.73 & 64.51 & 56.03 & 72.60 &  \textbf{73.07} \\ 
Error Bar & - & \suppstd{0.34} & \suppstd{0.16} & \suppstd{0.12} &  \suppstd{0.10} \\ 
\bottomrule
\end{tabular}
} 

\vspace{-.1in}
\end{table}

\subsection{Analyzing the Different Gradient Regularity Methods}

We explore the influence of gradient dropout regularity with 16 shots ImageNet in~\cref{sec:ab}.
of the manuscript paper. In this section, we provide more analysis of the different gradient regularity methods and mainly discuss the difference between GradSignDrop~\cite{chen2020just} and our R-AMT. The experimental results are shown in~\cref{table:graddrop_all}. We regard how to utilize the CE loss and KL loss as multi-task learning, with a key emphasis on balancing the general knowledge imparted by the KL loss and the specific knowledge captured by the CE loss. Given the low-data regime inherent in this setting, it is crucial to prevent overfitting in the CE loss and instead prioritize exploration to acquire specific knowledge while retaining the general knowledge present in the pre-trained weights (\ie, KL loss). Previous multi-task learning used gradient surgery to balance the different tasks, which does not consider the property of a low-data regime. Thus, directly applying the gradient surgery (\ie, AgreeGrad and GradSign) in the low-data regime does not bring performance improvement. Zhu \etal~\cite{zhu2022prompt} try to adapt PCGrad~\cite{yu2020gradient} to this task, which brings a slight 0.1\% performance improvement but is 0.37\% lower than R-AMT. We analyze that all conflict gradients forced to be projected in the vertical direction bring the overconfidence of general knowledge from KL loss. Our gradient dropout regularity does not change the direction of the CE gradient and provides a transformation of the gradient numerical scale, which can better explore the specific knowledge in the few-shot data regime.
In addition, R-AMT adds some level of randomness to the gradient guided by the KL divergence, which helps the model generalize better to downstream tasks.

 \begin{table}[htbp]
	\caption{\textbf{Ablation studies on different gradient regularity strategies.} The proposed gradient dropout regularity can make better use of general knowledge of KL loss while exploring the knowledge of downstream data.
 }
        \vspace{2pt}
	\centering\small
        \setlength{\tabcolsep}{6pt}
		\begin{tabular}{l|c|c}
		\toprule

            Method & Accuracy & Gain\\
            \hline 
            Zero-shot CLIP & 66.73 & -   \\
            \hline 
            AMT         & 72.60\std{0.12} & -   \\
            AMT+KL loss & 71.92\std{0.06} & \textcolor{green}{-0.68}  \\
            AMT+GradSign~\cite{chen2020just} & 71.95\std{0.08} & \textcolor{green}{-0.65} \\
            AMT+AgreeGrad~\cite{mansilla2021domain} & 68.82\std{0.09} & \textcolor{green}{-3.78}\\
            AMT+ProGrad~\cite{zhu2022prompt} & 72.70\std{0.22} & \textcolor{red}{+0.10} \\
            \hline
            R-AMT & \textbf{73.07}\std{0.10} & \textcolor{red}{+0.47} \\
            
		\bottomrule
		\end{tabular}
  \vspace{-.1in}
		\label{table:graddrop_all}
  \end{table}
  
\subsection{Applying Mask Tuning on Text/Image Encoders of CLIP}

\begin{figure*}[htbp]
    \centering
    \includegraphics[scale=0.28]{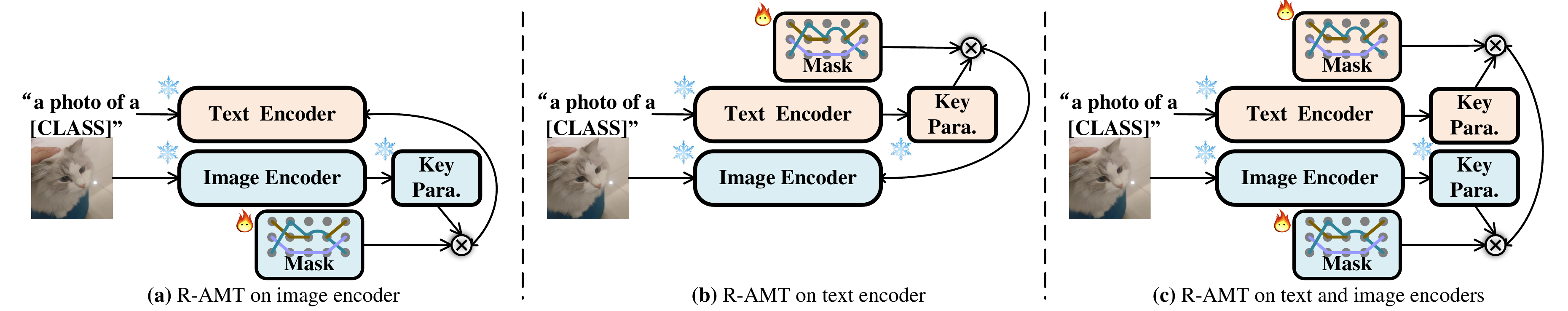}
    \caption{Applying the regularized mask tuning on different encoders of CLIP.
    }
    \label{fig:mulitimodel_mask}
\end{figure*}

To further validate the effectiveness of mask tuning on different encoders, we also adopt R-AMT on the text/image encoders of CLIP, as shown in~\cref{fig:mulitimodel_mask}. \cref{table:different_encoder} demonstrates that the performance of R-AMT on the image encoder is comparable to that of R-AMT on the text encoder, while requiring less training time. Notably, the best performance is achieved when R-AMT is applied to both the image and text encoders. Considering the balance between training time and performance, we have chosen to adopt R-AMT on the image encoder as our method.

\begin{table*}[htbp]
	\caption{Influence of applying the regularized mask tuning on different encoders of CLIP on 16-shot ImageNet.}
	
	\centering
	\resizebox{0.82\linewidth}{!}{
		\begin{tabular}{c|c|ccc}
		\toprule
		\multirow{2}{*}{Methods} & \multirow{2}{*}{Zero-shot CLIP} & \multicolumn{3}{c}{R-AMT}                                                                            \\ \cline{3-5} 
                         &                                 & \multicolumn{1}{c}{Image Encoder} & \multicolumn{1}{c}{Text Encoder} & Image Encoder + Text Encoder \\ 
		\hline 
	
	
		Accuracy & 66.73  &73.07& 73.05 & \textbf{74.00} \\
		Error Bar & - & \suppstd{0.10}& \suppstd{0.07} & \suppstd{0.14}  \\
            \hline
            Training Time (s/per image) & - & \textbf{0.07} & 1.04 & 1.67 \\
		\bottomrule
		\end{tabular}
		}    
		\label{table:different_encoder}
		\vspace{-.1in}
	\end{table*}

\subsection{Different Pruning-Based Mask Technologies}

\begin{table*}[htbp]
	\caption{Different parameter-level mask tuning on 16-shot ImageNet.}
	\centering
	\resizebox{0.82\linewidth}{!}{
		\begin{tabular}{c|c|c|c|c}
		\toprule
        \multirow{2}{*}{Methods} & \multirow{2}{*}{Zero-shot CLIP} & \multicolumn{3}{c}{R-AMT}\\ \cline{3-5} 
        & & \multicolumn{1}{c}{Filter-wise Pruning~\cite{he2019filter}} & \multicolumn{1}{c}{Channel-wise Pruning~\cite{liu2017learning}} & Parameter Pruning~\cite{yang2021nvit} \\ 
		\hline 
		Accuracy & 66.73 & 68.32 & 67.70 & \textbf{73.07}  \\
		Error Bar & -  & \suppstd{0.18} & \suppstd{0.27} & \suppstd{0.10} \\
		\bottomrule
		\end{tabular}
		} 
  \vspace{-.1in}
		\label{table:mask_channel}
		
\end{table*}

Recently, structured network pruning techniques~\cite{liebenwein2019provable,sui2021chip,gao2018dynamic} have been proposed to remove parameters in groups by pruning filters~\cite{he2019filter}, channels~\cite{liu2017learning}, or parameters~\cite{yang2021nvit}.
Inspired by these network pruning works, we adopt different pruning-based mask technologies from the dimension aspect, which are classified by Filter-wise Pruning, Channel-wise Pruning, and Parameter Pruning. 
Concretely, we adopt the channel-wise pruning method to the mask tuning method that focuses on the pruning of the input channel, while the filter-wise pruning method focuses on the pruning of the output channel. These two prompt learning are with all the details of dependencies reversed. 
As shown in~\cref{table:mask_channel}, Filter-wise Pruning and Channel-wise Pruning bring relatively low gains in accuracy compared to Zero-shot CLIP on 16-shot ImageNet. It likely neglects some important details in the pre-trained model when we just focus on measuring the importance of filter-wise or channel-wise information.
Parameter pruning results in the best performance, indicating that selecting more finely-tuned masks can enhance the search for more appropriate knowledge from pre-trained weights.

\subsection{Dynamic Mask Tuning}
We find the best performance of mask tuning on differernt datasets is achieved when we perform masking on different kinds of layers in~\cref{table:different_layes_vit}.
For example, R-AMT surpasses other methods on Caltech101 dataset, while R-MMT reaches the highest accuracy on StanfordCars dataset. R-AMT and R-MMT mean that we apply mask tuning on the MHSA and MLP layers, respectively.
Thus, we consider dynamically selecting layers to perform masking. We denote it Dynamic Mask Tuning (R-DMT).
Concretely, we aggregate the gradients from CE loss on each layer for one epoch before starting to train the mask. For each element in the leanable mask weight, positive gradient drives the element to be small, as shown in~\cref{eq:m_grad}. Once, the value of the element falls below the hard threshold $\alpha$, the corresponding binary mask becomes 0. Thus, we calculate the mean gradient for each layer and perform masking on the layer with positive mean gradient value. 
The experimental results are presented in~\cref{table:dynamic_layes_vit}.
We observe R-AMT surpasses R-DMT 0.35\% on the average across 11 datasets. The gradient of each element is changing during training period. Aggregating gradient before training to decide which layer to applying mask can not well unleash the potential of mask tuning. Thus, we choose performing mask tuning on MHSA layers.

\begin{table*}[htbp]
	\caption{Compare dynamically choosing layers with specifying different layers for performing masking on 16-shot datasets.}
        \vspace{2pt}
	\centering\small
        \setlength{\tabcolsep}{4.8pt}
		\begin{tabular}{c|ccccccccccccc}
		\toprule
		\makebox[0.05\textwidth][c]{\rotatebox{45}{Method}} &
		\makebox[0.05\textwidth][c]{\rotatebox{45}{ImageNet}} &
		\makebox[0.05\textwidth][c]{\rotatebox{45}{Caltech101}} &
		\makebox[0.05\textwidth][c]{\rotatebox{45}{FGVCAircraft}} &
		\makebox[0.05\textwidth][c]{\rotatebox{45}{StanfordCars}} & 
            \makebox[0.05\textwidth][c]{\rotatebox{45}{Flowers102}} & 
            \makebox[0.05\textwidth][c]{\rotatebox{45}{OxfordPets}} &
		\makebox[0.05\textwidth][c]{\rotatebox{45}{Food101}} &
		\makebox[0.05\textwidth][c]{\rotatebox{45}{DTD}} &
		\makebox[0.05\textwidth][c]{\rotatebox{45}{EuroSAT}} &
		\makebox[0.05\textwidth][c]{\rotatebox{45}{UCF101}} &
		\makebox[0.05\textwidth][c]{\rotatebox{45}{SUN397}} & 
		\makebox[0.05\textwidth][c]{\rotatebox{45}{Average}}\\
 
		\hline 
		\multirow{1}{*}{R-AMT} & 73.07 & \textbf{97.00} & 58.47 & 85.93 & 98.17 & 93.80 & 87.47 & 74.57 & \textbf{91.80} & 86.93 & \textbf{76.40} & \textbf{83.96}  \\
             Error Bar& \suppstd0.10 & \suppstd0.37 & \suppstd0.38 & \suppstd0.34 & \suppstd0.09 & \suppstd0.29 & \suppstd0.09 & \suppstd0.56 & \suppstd0.70 & \suppstd0.42 & \suppstd0.03 & -\\
             \hline
            \multirow{1}{*}{R-MMT} & \textbf{73.52} & 96.77 & 59.57 & \textbf{86.43} & 98.07 & 93.83 & 87.40 & \textbf{75.73} & 84.07 & \textbf{87.70} & 74.23 & 83.39  \\
          Error Bar& \suppstd0.15 & \suppstd0.39 & \suppstd0.05 & \suppstd0.09 & \suppstd0.05 & \suppstd0.38 & \suppstd0.16 & \suppstd0.39 & \suppstd1.02 & \suppstd0.16 & \suppstd0.05 & -\\
            \hline
            \multirow{1}{*}{R-PMT} & 73.48 & 96.63 & \textbf{60.30} & 86.33 & \textbf{98.27} & 93.77 & \textbf{87.50} & 75.60 & 88.20 & 87.33 & 76.12 & \textbf{83.96}  \\
           Error Bar& \suppstd0.11 & \suppstd0.29 & \suppstd0.82 & \suppstd0.17 & \suppstd0.12 & \suppstd0.25 & \suppstd0.08 & \suppstd0.51 & \suppstd4.69 & \suppstd0.26 &  \suppstd0.16 & -\\
            \hline
            \multirow{1}{*}{R-DMT} & 73.41 & 96.81 & 59.70 & 86.28 & 97.77 & 93.41 & 87.44 & 75.47 & 87.88 & 87.65 & 73.89 &  83.61\\
		Error Bar& \suppstd0.17 & \suppstd0.17 & \suppstd1.00 & \suppstd0.19 & \suppstd0.06 & \suppstd0.43 & \suppstd0.15 & \suppstd0.19 & \suppstd3.47 & \suppstd0.55 &  \suppstd0.11 & -\\
		\bottomrule
		\end{tabular}
		\label{table:dynamic_layes_vit}
		
        \vspace{-.1in}
	\end{table*}

\subsection{Base-to-new Generalization Results}
In~\cref{tab:b2n}, we demonstrate the numerical experimental results on each dataset on a 16-shot base-to-new generalization setting. The mask tuning methods (AMT and R-AMT) outperform other methods on 8 out of 11 datasets on the harmonic mean of accuracy on base and new classes. Moreover, we observe the gradient dropout regularity formalism significantly improves the harmonic mean of the accuracy of AMT on fine-grained classification tasks, \eg, StanfordCars, and tasks with a small amount of classes, \eg, EuroSAT. It indicates R-AMT is able to learn more reliable binary masks for fine-grained tasks than AMT. And R-AMT has an anti-overfitting ability, which improves the accuracy of mask tuning when the amount of training classes is limited.

\subsection{Few-Shot Recognition Accuracy}
The full numerical results of~\cref{fig:main_results_vit}
in the main text are presented in~\cref{tab:few_shot_results}. The highest accuracy in each shot setting and dataset are highlighted in red, while the second best is present in orange.
The original TIP-Adapter~\cite{zhang2022tip} utilizes prompt ensembling to construct text input on ImageNet, which provides better performance than a single prompt on Zero-shot CLIP.
Thus, we re-run TIP-Adapter with a single text prompt for a fair comparison. The comparison with TIP-Adapter when using prompt ensembling is presented in~\cref{sec:prompt_ensembling}. 
Overall, R-AMT achieves the best performance on the average of 11 datasets across all shot settings.

\section{Visualization}
\subsection{IoU of Masks among 11 Datasets.} 
\begin{figure}[htbp]
\centering
\subfloat[AMT]{
     \centering
     \includegraphics[width=0.45\linewidth]{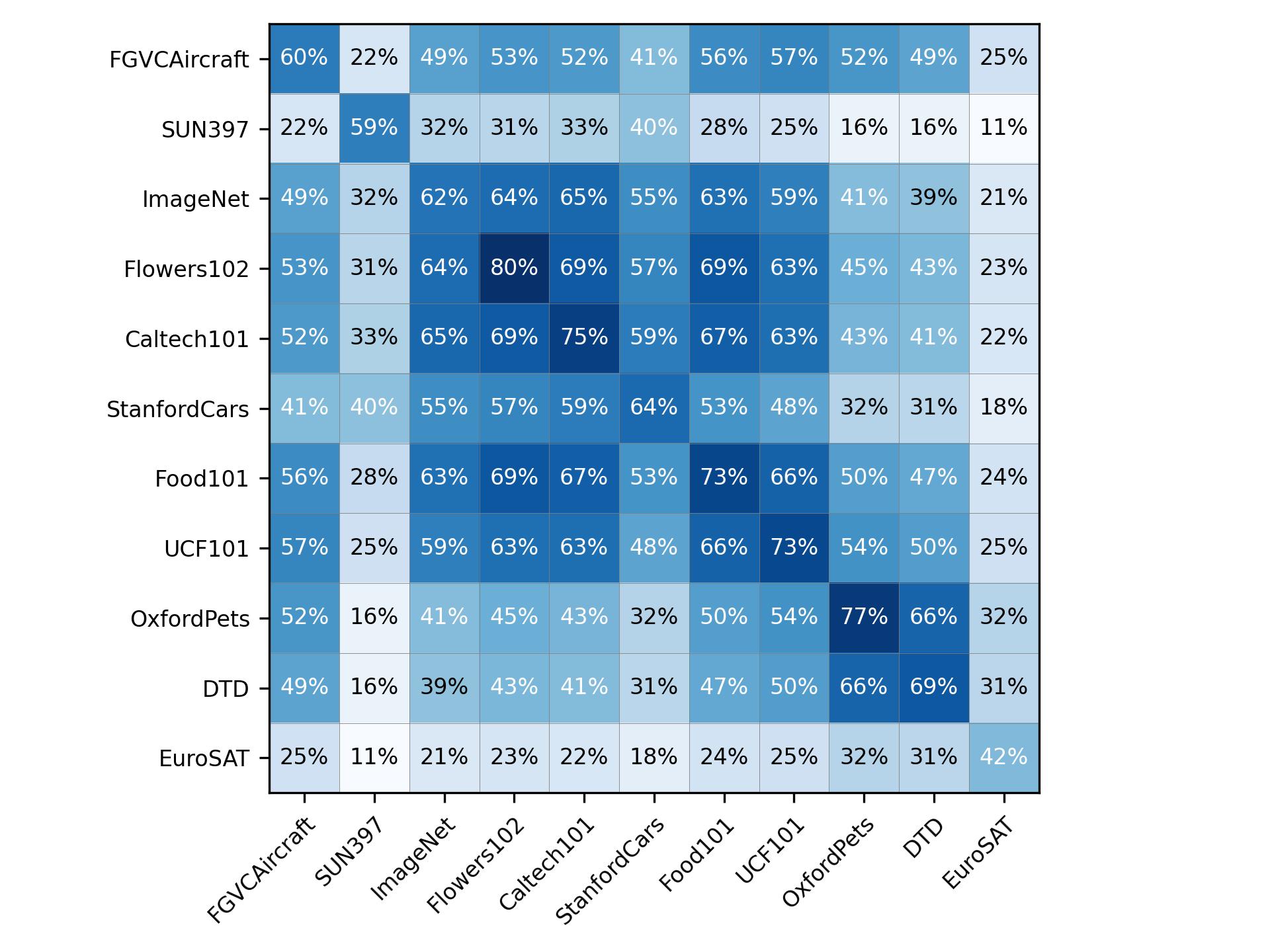}
     \label{fig:MASK_IoU}
 }
\subfloat[R-AMT]{
     \centering
     \includegraphics[width=0.45\linewidth]{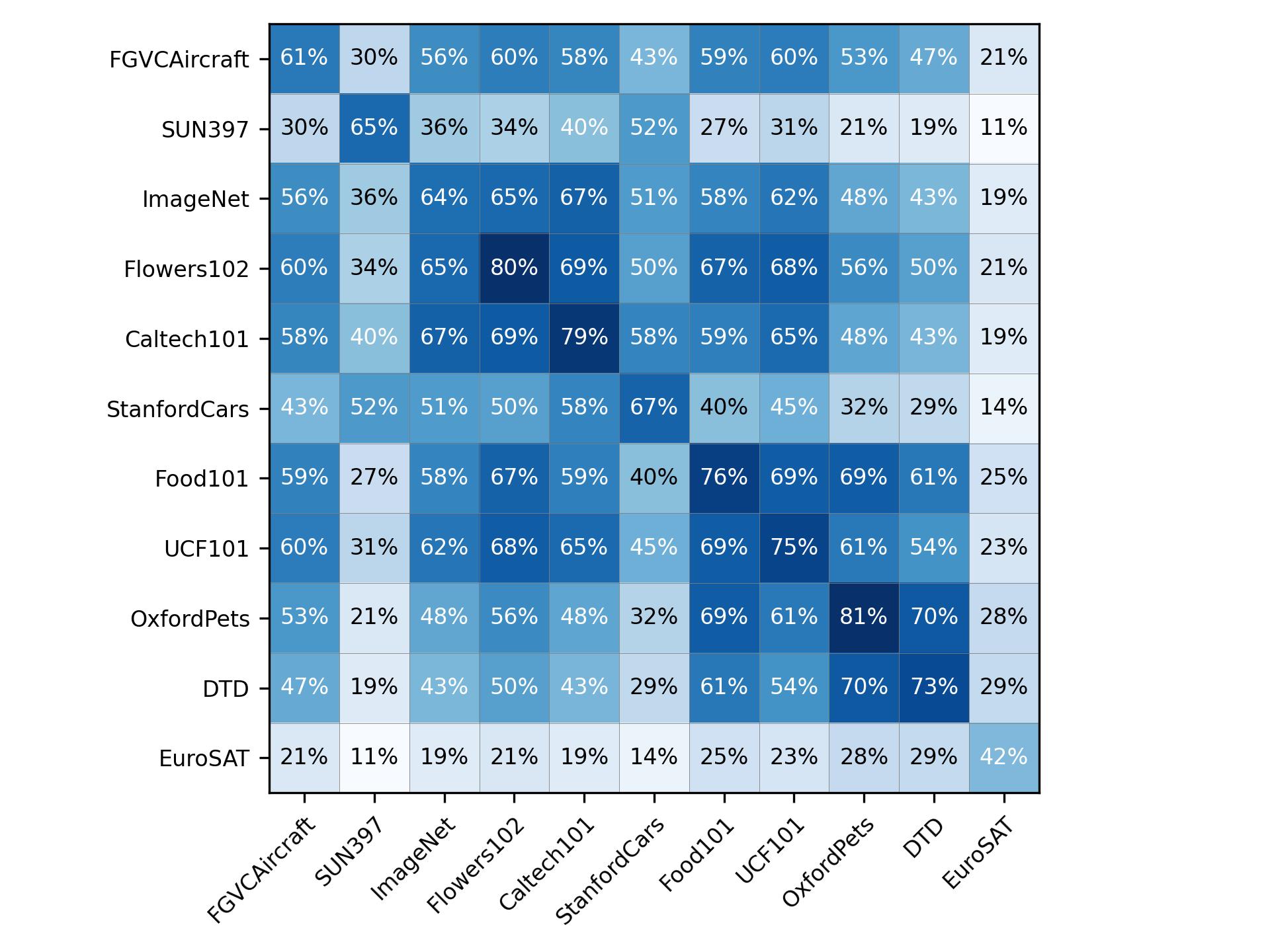}
     \label{fig:PROGRADMASK_IoU}
}
\caption{IoU between different binary masks among 11 datasets learned by AMT~(a) and R-AMT~(b).}
\label{fig:cm}
\end{figure}

As shown in~\cref{fig:cm}, we present the IoU of binary masks between two arbitrary datasets on the 16-shot setting. Since we random sample 16 images per class for training three times with different seeds, the binary masks within one dataset are not always the same. This result indicates the knowledge of pre-trained weight is not invariable for downstream classification tasks.
We observe that for each dataset the maximum IoU is always itself, which indicates the AMT and R-AMT can find task-specific parameters within CLIP. Moreover, the IoU of binay masks learned by R-AMT within one dataset is higher than AMT. It indicates the R-AMT is able to learn more stable binary masks in different runs.

  \begin{table*}[htbp]
    \centering
    \caption{Comparison on the base-to-new generalization setting with CoCoOP~\cite{zhou2022conditional}, ProGrad~\cite{zhu2022prompt} and CLIP-adapter~\cite{gao2021clip} with 16-shots. H denotes the harmonic mean of the accuracy on base and new classes. All methods are trained on the base classes. We report the average results and standard deviation over three runs for AMT and R-AMT. }
    \label{tab:b2n}
    \begin{subfloat}[\textbf{Average over 11 datasets}]{
    \centering
    \resizebox{2.15in}{!}{
    \begin{tabular}{ccc|c}
    \toprule
         & Base & New & H \\
         \hline
       Zero-shot CLIP  &  69.34 & \textbf{74.22} & 71.70 \\
       CoCoOP  & 80.47 & 71.69 & 75.83 \\
       ProGrad  & 82.79 & 68.55 & 75.00 \\
       CLIP-adapter  & 82.62 & 70.97 & 76.35 \\
       \hline
       \multirow{2}{*}{AMT}  & \textbf{86.17} & 69.11 & 76.70  \\
       & - & - &  - \\
       \hline
       \multirow{2}{*}{R-AMT} & 85.71 &72.15 & \textbf{78.35} \\
       & - & - & - \\
    \bottomrule
    \end{tabular}}
    \label{tab:Average}}
    \end{subfloat}
     \hfill
    \begin{subfloat}[FGVCAircraft]{
    \centering
    \resizebox{2.2in}{!}{
    \begin{tabular}{ccc|c}
    \toprule
         & Base & New & H \\
         \hline
       Zero-shot CLIP  &  27.19& \textbf{36.29} & 31.09 \\
       CoCoOP  & 33.41 & 23.71 & 27.74 \\
       ProGrad  & 42.63 & 26.97 & 33.04 \\
       CLIP-adapter  & 39.57 & 32.27 & 35.55 \\
       \hline
       \multirow{2}{*}{AMT}  &\textbf{52.42}  & 28.11 & 36.60 \\
       & \suppstd{0.85} & \suppstd{0.75} & - \\
       \hline
       \multirow{2}{*}{R-AMT} & 49.22 & 32.09 & \textbf{38.85} \\
       & \suppstd{0.68} & \suppstd{1.11} & - \\
    \bottomrule
    \end{tabular}}
    \label{tab:FGVCAircraft}}
    \end{subfloat}
     \hfill
    \begin{subfloat}[ImageNet]{
    \centering
    \resizebox{2.2in}{!}{
    \begin{tabular}{ccc|c}
    \toprule
         & Base & New & H \\
         \hline
       Zero-shot CLIP  &  72.43& 68.14 &70.22 \\
       CoCoOP  & 75.98 &\textbf{70.43} &73.10 \\
       ProGrad  & 77.03 &68.80 &72.68 \\
       CLIP-adapter  & 76.53 &66.67& 71.26 \\
       \hline
       \multirow{2}{*}{AMT}  & \textbf{77.23} & 70.30 & \textbf{73.60} \\
       & \suppstd{0.07} & \suppstd{0.24} & - \\
       \hline
       \multirow{2}{*}{R-AMT} & 77.22 & 70.28 & 73.59\\
       & \suppstd{0.17} & \suppstd{0.02} & - \\
    \bottomrule
    \end{tabular}}
    \label{tab:ImageNet}}
    \end{subfloat}

    \begin{subfloat}[StanfordCars]{
    \centering
    \resizebox{2.2in}{!}{
    \begin{tabular}{ccc|c}
    \toprule
         & Base & New & H \\
         \hline
       Zero-shot CLIP  &  63.37 &\textbf{74.89}& 68.65\\
       CoCoOP  & 70.49 &73.59 &72.01 \\
       ProGrad  & 79.00 &67.93& 73.05 \\
       CLIP-adapter  & 77.13 &69.23 &72.97 \\
       \hline
       \multirow{2}{*}{AMT}  & 83.49 & 62.52 & 71.50\\
       & \suppstd{0.44} & \suppstd{0.50} & - \\
       \hline
       \multirow{2}{*}{R-AMT} & \textbf{82.90} & 69.46& \textbf{75.59} \\ 
       & \suppstd{0.21} & \suppstd{0.49} & - \\
   \bottomrule
    \end{tabular}}
    \label{tab:StanfordCars}}
    \end{subfloat}
    \hfill
    \begin{subfloat}[Caltech101]{
    \centering
    \resizebox{2.2in}{!}{
    \begin{tabular}{ccc|c}
    \toprule
         & Base & New & H \\
         \hline
       Zero-shot CLIP  &  96.84 & 94.00 &95.40 \\
       CoCoOP  & 97.96 &93.81 &95.84 \\
       ProGrad  & 98.50 &91.90 &95.09 \\
       CLIP-adapter  & 98.20& 93.20& 95.63 \\
       \hline
       \multirow{2}{*}{AMT}  & \textbf{98.88} & \textbf{94.61} & \textbf{96.70} \\
       & \suppstd{0.16} & \suppstd{0.27} & - \\
       \hline
       \multirow{2}{*}{R-AMT} & 98.88 & 94.43 & 96.60\\
       & \suppstd{0.21} & \suppstd{0.16} & - \\
    \bottomrule
    \end{tabular}}
    \label{tab:Caltech101}}
    \end{subfloat}
    \hfill
    \begin{subfloat}[UCF101]{
    \centering
    \resizebox{2.2in}{!}{
    \begin{tabular}{ccc|c}
    \toprule
         & Base & New & H \\
         \hline
       Zero-shot CLIP  &  70.53& \textbf{77.50} &73.85 \\
       CoCoOP  & 82.33& 73.45& 77.64 \\
       ProGrad  & 83.90& 68.50 &75.42 \\
       CLIP-adapter  & 85.80 &73.63 &79.25 \\
       \hline
       \multirow{2}{*}{AMT}  & \textbf{88.95} & 76.22 & 82.09 \\
       & \suppstd{0.41} & \suppstd{0.55} & - \\
       \hline
       \multirow{2}{*}{R-AMT} & 87.87 & 77.39 & \textbf{82.30} \\
        & \suppstd{0.38} & \suppstd{0.67} & - \\
    \bottomrule
    \end{tabular}}
    \label{tab:UCF101}}
    \end{subfloat}

    \begin{subfloat}[EuroSAT]{
    \centering
    \resizebox{2.2in}{!}{
    \begin{tabular}{ccc|c}
    \toprule
         & Base & New & H \\
         \hline
       Zero-shot CLIP  &  56.48 &64.05 &60.03 \\
       CoCoOP  & 87.49& 60.04 &71.21 \\
       ProGrad  & 91.37& 56.53& 69.85 \\
       CLIP-adapter  & 86.93& \textbf{64.20}& \textbf{73.86} \\
        \hline
       \multirow{2}{*}{AMT}  & \textbf{97.01} & 51.61 & 67.38 \\
       & \suppstd{0.81} & \suppstd{4.06} & - \\
       \hline
       \multirow{2}{*}{R-AMT} & 95.79 & 58.25 & 72.45\\
       & \suppstd{1.77} & \suppstd{5.38} & - \\
    \bottomrule
    \end{tabular}}
    \label{tab:EuroSAT}}
    \end{subfloat}
    \hfill
    \begin{subfloat}[Flowers102]{
    \centering
    \resizebox{2.2in}{!}{
    \begin{tabular}{ccc|c}
    \toprule
         & Base & New & H \\
         \hline
       Zero-shot CLIP  &  72.08 & \textbf{77.80}& 74.83 \\
       CoCoOP  & 94.87 &71.75 &81.71 \\
       ProGrad  & 96.27& 71.07 &81.77 \\
       CLIP-adapter  & 97.70 &70.83& 82.13 \\
        \hline
       \multirow{2}{*}{AMT}  & \textbf{98.32} & 65.13 & 78.36 \\
       & \suppstd{0.05} & \suppstd{1.34} & - \\
       \hline
       \multirow{2}{*}{R-AMT} & 97.95 & 70.90 & \textbf{82.26} \\
       & \suppstd{0.09} & \suppstd{1.48} & - \\
    \bottomrule
    \end{tabular}}
    \label{tab:Flowers102}}
    \end{subfloat}
    \hfill
    \begin{subfloat}[Food101]{
    \centering
    \resizebox{2.2in}{!}{
    \begin{tabular}{ccc|c}
    \toprule
         & Base & New & H \\
         \hline
       Zero-shot CLIP  &  90.10 &91.22 &90.66 \\
       CoCoOP  & \textbf{90.70} &\textbf{91.29} &\textbf{90.99} \\
       ProGrad  & 90.17 &89.53 &89.85 \\
       CLIP-adapter  & 90.40& 90.40& 90.40 \\
       \hline
       \multirow{2}{*}{AMT}  & 89.81 &90.26 & 90.03\\
       & \suppstd{0.08} & \suppstd{0.33} & - \\
       \hline
       \multirow{2}{*}{R-AMT} & 90.69 & 91.14 & 90.91\\
       & \suppstd{0.10} & \suppstd{0.24} & - \\
    \bottomrule
    \end{tabular}}
    \label{tab:Food101}}
    \end{subfloat}

    \begin{subfloat}[SUN397]{
    \centering
    \resizebox{2.2in}{!}{
    \begin{tabular}{ccc|c}
    \toprule
         & Base & New & H \\
         \hline
       Zero-shot CLIP  &  69.36 &75.35& 72.23 \\
       CoCoOP  & 79.74 &\textbf{76.86} &78.27 \\
       ProGrad  & 80.70& 71.03 &75.56 \\
       CLIP-adapter  & 81.67 &73.93 &77.61 \\
        \hline
       \multirow{2}{*}{AMT}  & 80.99 & 72.81 & 76.68 \\
       & \suppstd{0.31} & \suppstd{0.30} & - \\
       \hline
       \multirow{2}{*}{R-AMT} & \textbf{82.15} & 76.53 &\textbf{79.24} \\
       & \suppstd{0.23} & \suppstd{0.25} & - \\
    \bottomrule
    \end{tabular}}
    \label{tab:SUN397}}
    \end{subfloat}
    \hfill
    \begin{subfloat}[OxfordPets]{
    \centering
    \resizebox{2.2in}{!}{
    \begin{tabular}{ccc|c}
    \toprule
         & Base & New & H \\
         \hline
       Zero-shot CLIP  &  91.17& \textbf{97.26}& 94.12 \\
       CoCoOP  & 95.20 &97.69& \textbf{96.43} \\
       ProGrad  & 94.40& 95.10& 94.75 \\
       CLIP-adapter  & 94.40& 94.10& 94.25 \\
        \hline
       \multirow{2}{*}{AMT}  & 95.53 & 96.14 & 95.83 \\
       & \suppstd{0.27} & \suppstd{0.96} & - \\
       \hline
       \multirow{2}{*}{R-AMT} & \textbf{95.68} & 96.01 & 95.84 \\
       & \suppstd{0.24} & \suppstd{1.02} & - \\
    \bottomrule
    \end{tabular}}
    \label{tab:OxfordPets}}
    \end{subfloat}
    \hfill
    \begin{subfloat}[DTD]{
    \centering
    \resizebox{2.2in}{!}{
    \begin{tabular}{ccc|c}
    \toprule
         & Base & New & H \\
         \hline
       Zero-shot CLIP  &  53.24& \textbf{59.90} &56.37 \\
       CoCoOP  & 77.01 &56.00& 64.85 \\
       ProGrad  & 76.70 &46.67 &58.03 \\
       CLIP-adapter  & 80.47& 52.23& 63.35 \\
        \hline
       \multirow{2}{*}{AMT}  & \textbf{85.26} & 52.54 & 65.02 \\
       & \suppstd{0.48} & \suppstd{1.23} & - \\
       \hline
       \multirow{2}{*}{R-AMT} & 84.41 & 57.17 & \textbf{68.17} \\
       & \suppstd{0.52} & \suppstd{0.88} & - \\
    \bottomrule
    \end{tabular}}
    \label{tab:DTD}}
    \end{subfloat}
    
    
\end{table*}

\begin{table}[htbp]
  \centering
  \footnotesize
  \caption{\textbf{Accuracy (\%) of few-shot learning, \ie, 16/8/4/2/1-shot, on the 11 datasets.} We report the average accuracy over three runs. ``F.A.'' refers to FGVCAircraft, ``S.C.'' refers to StanfordCars.}
  \setlength{\tabcolsep}{3pt}
    \begin{tabular}{c|c|c|c|c|c|c|c|c|c|c|c|c|c}
    \toprule
    shot & Method & F.A. & ImageNet & OxfordPet & Flowers102 & EuroSAT & S.C. & Caltech101 & UCF101 & Food101 & SUN397 & DTD & Average \\
    \hline
    -     & Zero Shot & 24.72  & 66.73  & 89.21  & 71.34  & 47.60  & 65.32  & 92.94  & 66.75  & 86.06  & 62.50  & 44.39  & 65.23  \\
    \hline
    16    & Linear Prob & 36.45  & 56.03  & 76.40  & 94.91  & 82.67  & 70.01  & 90.72  & 73.72  & 70.80  & 67.15  & 63.42  & 71.12  \\
    \hline
    16    & CoOP  & 43.29  & 72.01  & 91.92  & 96.93  & 86.05  & 82.91  & 95.47  & 82.25  & 84.33  & 74.58  & 69.21  & 79.90  \\
    \hline
    16    & TIP-Adapter & 45.20  & \bm{\best{73.08}} & 92.66  & 96.15  & 88.53  & 83.04  & 95.63  & 84.24  &\bm{\seco{87.31 }} &\bm{\seco{76.21 }} & 71.57  & 81.24  \\
    \hline
    16    & ProGrad & 40.50  & 72.25  & 92.76  & 94.98  & 84.51  & 81.48  & 95.87  & 81.54  & 86.76  & 75.02  & 65.62  & 79.21  \\
    \hline
    16    & VPT-deep & 40.96  & 70.57  & 92.91  & 94.96  & 91.53  & 76.13  & 95.83  & 82.76  & 86.18  & 71.63  & 69.79  & 79.39  \\
    \hline
    16    & UPT   & 46.80  & 72.63  & 92.95  & 97.11  & 90.51  & 84.33  & 95.94  & 84.03  & 85.00  & 75.92  & 70.65  & 81.44  \\
    \hline
    16    & AMT   & \bm{\best{59.43 }} & 72.60  &\bm{\seco{93.43 }} &\bm{\seco{98.07 }} & \bm{\best{92.00 }} &\bm{\seco{85.70 }} & \bm{\best{97.10 }} & \bm{\best{87.00 }} & 85.93  & 72.27  &\bm{\seco{74.53 }} &\bm{\seco{83.46 }} \\
    16    & Error Bar   &\std0.58  &\std 0.12  &\std 0.48  &\std 0.17  &\std 0.75  &\std 0.36  &\std 0.22  &\std 0.62  &\std 0.09  &\std 0.21  &\std 0.25 & -  \\
    \hline
        16    & R-AMT &\bm{\seco{58.47 }} &\bm{\seco{73.07 }} & \bm{\best{93.80 }} & \bm{\best{98.17 }} &\bm{\seco{91.80 }} & \bm{\best{85.93 }} &\bm{\seco{97.00 }} &\bm{\seco{86.93 }} & \bm{\best{87.47 }} & \bm{\best{76.40 }} & \bm{\best{74.57 }} & \bm{\best{83.96 }} \\
    16    & Error Bar   &\std0.38  &\std 0.10  &\std 0.29  &\std 0.09  &\std 0.70  &\std 0.34  &\std 0.37  &\std 0.42  &\std 0.09  &\std 0.03  &\std 0.56 & -  \\
    \hline
    \midrule
    shot & Method & F.A. & ImageNet & OxfordPet & Flowers102 & EuroSAT & S.C. & Caltech101 & UCF101 & Food101 & SUN397 & DTD & Average \\
    \hline
    -     & Zero Shot & 24.72  & 66.73  & 89.21  & 71.34  & 47.60  & 65.32  & 92.94  & 66.75  & 86.06  & 62.50  & 44.39  & 65.23  \\
    \hline
    8     & Linear Prob & 29.46  & 49.67  & 66.36  & 92.03  & 77.58  & 60.90  & 88.03  & 69.47  & 63.99  & 62.24  & 57.15  & 65.17  \\
    \hline
    8     & CoOP  & 39.16  & 70.68  & 91.62  & 94.92  & 78.71  & 78.79  & 94.46  & 80.02  & 82.66  & 71.36  & 65.01  & 77.04  \\
    \hline
    8     & TIP-Adapter & 40.79  & \bm{\seco{71.42 }} & 91.75  & 93.94  &\bm{\seco{83.23}}  & 78.46  & 95.36  & 82.03  & \bm{\seco{86.78 }} & \bm{\seco{73.44 }} & 66.31  & 78.50  \\
    \hline
    8     & ProGrad & 37.70  & 71.06  & 92.12  & 93.49  & 79.29  & 78.75  & 94.92  & 79.64  & 85.77  & 72.84  & 62.35  & 77.08  \\
    \hline
    8     & VPT-deep & 36.38  & 69.83  & 92.28  & 91.53  & 80.75  & 72.61  & 95.37  & 80.16  & 85.20  & 69.90  & 64.06  & 76.19  \\
    \hline
    8     & UPT   & 39.69  & 71.60  & 92.78  & 95.32  & \bm{\best{85.53} } & 79.95  & 95.04  & 80.93  & 86.14  & 74.00  & 65.57  & 78.78  \\
    \hline
    8     & AMT   & \bm{\best{47.40 }} & 70.33  & \bm{\seco{92.47 }} & \bm{\best{96.47 }} & 82.00  & \bm{\best{80.23 }} & \bm{\best{96.30 }} & \bm{\best{85.00 }} & 85.07  & 68.30 & \bm{\best{71.30 }} & \bm{\seco{79.53 }} \\
    8     &  Error Bar     &\std 0.67  &\std 0.34  &\std 0.19  &\std 0.62  &\std 0.97  &\std 0.12  &\std 0.28  &\std 0.57  &\std 0.17  &\std 0.42  &\std 0.37 &- \\
    \hline
    8     & R-AMT & \bm{\seco{45.40 }} & \bm{\best{71.50 }} & \bm{\best{93.63 }} & \bm{\seco{95.57 }} & 82.53 & \bm{\seco{80.97 }} & \bm{\seco{96.10 }} & \bm{\seco{84.57 }} & \bm{\best{87.13 }} & \bm{\best{73.47 }} & \bm{\seco{70.20 }} & \bm{\best{80.10 }} \\
    8     &  Error Bar       &\std 0.67  &\std 0.28  &\std 0.19  &\std 0.62  &\std 0.97  &\std 0.12  &\std 0.28  &\std 0.57  &\std 0.17  &\std 0.25  &\std 0.37 & -\\
    \hline     \midrule
    shot & Method & F.A. & ImageNet & OxfordPet & Flowers102 & EuroSAT & S.C. & Caltech101 & UCF101 & Food101 & SUN397 & DTD & Average \\
    \hline
    -     & Zero Shot & 24.72  & 66.73  & 89.21  & 71.34  & 47.60  & 65.32  & 92.94  & 66.75  & 86.06  & 62.50  & 44.39  & 65.23  \\
    \hline
    4     & Linear Prob & 23.70  & 41.51  & 56.09  & 84.84  & 69.39  & 48.52  & 82.95  & 62.32  & 55.11  & 54.61  & 50.08  & 57.19  \\
    \hline
    4     & CoOP  & 31.23  & 68.91  & 92.23 & 91.93  & 72.12  & 74.50  & 94.43  & 76.96  & 84.35  & 69.70  & 59.85  & 74.20  \\
    \hline
    4     & TIP-Adapter & 34.90 & 69.83 & 91.53  & 90.74  & \bm{\seco{77.91}} & 74.89  & 94.76  & 79.14  & \bm{\seco{86.53 }} & 70.22  & 61.96  & 75.67  \\
    \hline
    4     & ProGrad & 33.70  & 69.35  & 92.10  & 91.19  & 71.07  & 75.33 & 93.99  & 77.64  & 84.95  & 70.70 & 58.69  & 74.43  \\
    \hline
    4     & VPT-deep & 32.99  & 69.37  & \bm{\seco{92.40}} & 85.49  & 70.87  & 69.92  & 94.73  & 77.14  & 84.92 & 68.55  & 56.08  & 72.95  \\
    \hline
    4     & UPT   & 33.39  & \bm{\seco{70.28}}  & 92.10  & 92.11  & 75.17  & \bm{\seco{75.71}} & 94.09  & 77.53  & 85.34 & \bm{\seco{72.10}} & 60.87  & 75.34  \\
    \hline
    4     & AMT   & \bm{\best{37.80 }} & 69.93 & 92.03  & \bm{\best{93.87 }} & 72.23  & 75.03  & \bm{\best{96.40 }} & \bm{\best{81.87 }} & 84.73  & 70.80  & \bm{\best{65.47 }} & \bm{\seco{76.38 }} \\
    4     & Error Bar &\std 0.22  &\std 0.17  &\std 0.45  &\std 0.68  &\std 2.85  &\std 0.58  &\std 0.29  &\std 0.37  &\std 0.25  &\std 0.29  &\std 1.19  & - \\
    \hline
    4     & R-AMT & \bm{\seco{37.33}} & \bm{\best{70.80 }} & \bm{\best{92.80}} &  \bm{\seco{92.80}} & \bm{\best{81.87}} & \bm{\best{76.33}} & \bm{\seco{95.63}} & \bm{\seco{81.60}} & \bm{\best{86.63}} & \bm{\best{72.37 }} & \bm{\seco{65.27}}  & \bm{\best{77.58 }} \\
    4     & Error Bar &\std 0.19  &\std 0.16  &\std 0.14  &\std 0.37  &\std 1.47  &\std 0.66  &\std 0.19  &\std 0.22  &\std 0.05  &\std 0.37  &\std 1.54  & - \\
    \hline     \midrule
    shot & Method & F.A. & ImageNet & OxfordPet & Flowers102 & EuroSAT & S.C. & Caltech101 & UCF101 & Food101 & SUN397 & DTD & Average \\
    \hline
    -     & Zero Shot & 24.72  & 66.73  & 89.21  & 71.34  & 47.60  & 65.32  & 92.94  & 66.75  & 86.06  & 62.50  & 44.39  & 65.23  \\
    \hline
    2     & Linear Prob & 17.83  & 31.51  & 43.55  & 73.38  & 61.74  & 36.72  & 78.43  & 53.54  & 41.89  & 44.46  & 39.46  & 47.50  \\
    \hline
    2     & CoOP  & 26.85  & 66.71  & 90.07  & 87.63  & 64.71  & 70.88  & 92.70  & 74.03  & 84.38  & 66.98  & 53.86  & 70.80  \\
    \hline
    2     & TIP-Adapter & \bm{\best{32.78}}  &68.58  & \bm{\seco{91.10 }} & \bm{\best{90.49}} & \bm{\best{71.57}} & 70.07  & 93.68  & 76.09  & \bm{\best{86.29}} & 66.79  & 56.11  & \bm{\seco{73.05}}  \\
    \hline
    2     & ProGrad & 30.91  & 66.56  & 90.45  & 88.59  & 66.08  & \bm{\seco{71.62 }} & 93.09  & 74.30  & 84.27  & 68.28  & 54.63  & 71.71  \\
    \hline
    2     & VPT-deep & 29.36  & 68.64  & 90.50  & 77.60  & 69.28  & 68.03  & {94.70 } & 73.99  & 84.69  & 67.55  & 48.38  & 70.25  \\
    \hline
    2     & UPT   & 30.00  & \bm{\seco{69.90}}  & \bm{\best{92.50 }} & 81.88  & 68.96  & 69.44  & 94.17  & 74.89  & 85.02  & 69.75  & 52.98  & 71.77  \\
    \hline
    2     & AMT   & 30.46  & {69.28 } & 89.34  & \bm{\seco{88.86 }} & \bm{\seco{70.12 }} & 69.22  & \bm{\seco{94.48}}  & \bm{\best{78.46 }} & 84.38  & \bm{\seco{69.90 }} & \bm{\best{56.54 }} & 72.82  \\
    2     & Error Bar &\std 0.54  &\std 0.11  &\std 0.68  &\std 1.29  &\std 2.84  &\std 0.38  &\std 0.39  &\std 0.76  &\std 0.61  &\std 0.41  &\std 1.86  & - \\
    \hline
    2     & R-AMT & \bm{\seco{31.72 }} & \bm{\best{69.92}} & 90.82  & 88.41  & 69.02  & \bm{\best{72.46}} & \bm{\best{94.61} } & \bm{\seco{77.75 }} & \bm{\seco{86.26 }} &\bm{\best{71.00}} & \bm{\seco{56.32 }} & \bm{\best{73.48 }} \\
    2     & Error Bar &\std 0.32  &\std 0.16  &\std 0.45  &\std 1.68  &\std 2.61  &\std 0.30  &\std 0.37  &\std 0.61  &\std 0.25  &\std 0.22  &\std 1.72  & - \\
    \hline     \midrule
    shot & Method & F.A. & ImageNet & OxfordPet & Flowers102 & EuroSAT & S.C. & Caltech101 & UCF101 & Food101 & SUN397 & DTD & Average \\
    \hline
    -     & Zero Shot & 24.72  & 66.73  & 89.21  & 71.34  & 47.60  & 65.32  & 92.94  & 66.75  & 86.06  & 62.50  & 44.39  & 65.23  \\
    \hline
    1     & Linear Prob & 12.88  & 22.11  & 30.04  & 58.15  & 50.21  & 24.61  & 70.40  & 41.31  & 30.13  & 32.58  & 29.65  & 36.55  \\
    \hline
    1     & CoOP  & 21.33  & 65.82  & 90.40  & 78.89  & 53.62  & 67.36  & 93.06  & 71.50  & 84.29 & 67.05  & 50.91  & 67.66  \\
    \hline
    1     & TIP-Adapter & \bm{\seco{29.44} } & {67.41 } & \bm{\seco{90.79 }} & \bm{\best{86.26 }} & 63.92  & \bm{\seco{67.80 }} & {93.34 } & 73.38  & \bm{\best{86.13}} & 64.06  & \bm{\best{53.17 }} & \bm{\best{70.52}}  \\
    \hline
    1     & ProGrad & 27.95  & 64.40  & 88.94  & \bm{\seco{83.63 }} & 55.04  & 67.08  & 90.96  & 71.84  & 82.68  & 64.51  & \bm{\seco{52.74 }} & 68.16  \\
    \hline
    1     & VPT-deep & 28.23  & 68.28  & 90.44  & 71.95  & \bm{\best{66.89 }} & 66.68  & 93.06  & 71.03  & 84.15  & 66.70  & 45.38  & 68.44  \\
    \hline
    1     & UPT   & 28.47  & \bm{\best{69.68 }} & \bm{\best{92.04 }} & 74.67  & \bm{\seco{66.41 }} & 67.56  & {93.66}  & 71.93  & 84.10  & 68.85  & 45.09  & 69.31  \\
    \hline
    1     & AMT   & 28.94  & 68.98  & 89.46  & {83.46 } & 58.80  & 66.61  & \bm{\seco{93.75} } & \bm{\best{74.31} } & 83.97  & \bm{\seco{68.15 }} & 50.71  & \bm{\seco{69.74} } \\
    1     & Error Bar &\std 0.24  &\std 0.20  &\std 0.84  &\std 0.87  &\std 5.27  &\std 0.17  &\std 0.38  &\std 0.49  &\std 0.57  &\std 0.43  &\std 0.96  & - \\
    \hline
    1     & R-AMT & \bm{\best{29.47 }} & \bm{\seco{69.35 }} & 89.69  & {83.14 } & 61.03  & \bm{\best{69.30}} & \bm{\best{94.15 }} & \bm{\seco{74.08 }} & \bm{\seco{85.12 }} & \bm{\best{69.13} } & 51.28  & \bm{\best{70.52} } \\
    1     & Error Bar &\std 0.18  &\std 0.18  &\std 0.65  &\std 0.51  &\std 1.82  &\std 0.28  &\std 0.50  &\std 0.25  &\std 0.38  &\std 0.22  &\std 1.32  & - \\
    \bottomrule
    \end{tabular}%
  \label{tab:few_shot_results}%
\end{table}%

\clearpage